\documentclass[journal,12pt,onecolumn,draftclsnofoot]{IEEEtran}

\usepackage{color,array,amsthm}
\usepackage{tabularx}
\usepackage{graphicx}
\usepackage{authblk}
\usepackage{amsmath,amssymb,amsfonts}
\usepackage{algorithmic}
\usepackage{graphicx}
\usepackage{textcomp}
\usepackage{xcolor}
\usepackage{arydshln}
\usepackage{setspace}% http://ctan.org/pkg/setspace
\usepackage{lipsum}% http://ctan.org/pkg/lipsum
\usepackage{textcomp}
\usepackage{indentfirst}
\usepackage{accents}
\usepackage{soul}
\setuldepth{we}
\usepackage{enumitem}
\usepackage[font=footnotesize,labelfont=bf]{caption}
\newlength{\fudgeheight}

\usepackage{hyperref,tablefootnote,footnotehyper}
\newcommand{\thickhat}[1]{\mathbf{\hat{\text{$#1$}}}}

\newcommand{\thickdot}[1]{\mathbf{\dot{\text{$#1$}}}}
\newcommand{\thicktilde}[1]{\mathbf{\tilde{\text{$#1$}}}}
\newcommand{\thickbreve}[1]{\mathbf{\breve{\text{$#1$}}}}

\newcommand{\bbC}{{\mathbb{C}}}
\newcolumntype{a}{>{\hsize=1.3\hsize}X}
\newcolumntype{s}{>{\hsize=0.7\hsize}X}
\newcolumntype{P}[1]{>{\centering\arraybackslash}p{#1}}
\newcolumntype{L}[1]{>{\arraybackslash}p{#1}}
\DeclareMathOperator{\Tr}{Tr}
\renewcommand{\thesection}{\Roman{section}}
\renewcommand{\thesubsection}{\Alph{subsection}}

\makeatletter
\renewcommand{\p@subsection}{\thesection.}
\renewcommand{\p@subsubsection}{\thesection.\thesubsection.}
\makeatother

\begin{document}
\bstctlcite{IEEEexample:BSTcontrol}

\title{\LARGE Data-Driven Target Localization Using Adaptive Radar \\ \vspace{-1ex}Processing and Convolutional Neural Networks}

\author[*]{\vspace{1ex}\normalsize Shyam Venkatasubramanian} 
\author[$\dagger$]{\normalsize Sandeep Gogineni}
\author[$\ddagger$]{\normalsize Bosung Kang\vspace{-1ex}}
\author[$\S$]{\normalsize \\Ali Pezeshki}
\author[$\P$]{\normalsize Muralidhar Rangaswamy}
\author[*]{\normalsize Vahid Tarokh}

\affil[*]{\fontsize{10.7}{13}\selectfont Duke University, Durham, NC 27705, USA\vspace{-1ex}}
\affil[$\dagger$]{\fontsize{10.7}{13}\selectfont Information Systems Laboratories Inc., Dayton, OH 45431, USA\vspace{-1ex}}
\affil[$\ddagger$]{\fontsize{10.7}{13}\selectfont University of Dayton Research Institute, Dayton, OH 45469, USA\vspace{-1ex}}
\affil[$\S$]{\fontsize{10.7}{13}\selectfont Colorado State University, Fort Collins, CO 80523, USA\vspace{-1ex}}
\affil[$\P$]{\fontsize{10.7}{13}\selectfont AFRL, Wright-Patterson Air Force Base, OH 45433, USA\vspace{-4ex}}

\maketitle

\begin{spacing}{1.3}
\begin{abstract} Leveraging the advanced functionalities of modern radio frequency (RF) modeling and simulation tools, specifically designed for adaptive radar processing applications, this paper presents a data-driven approach to improve accuracy in radar target localization post adaptive radar detection. To this end, we generate a large number of radar returns by randomly placing targets of variable strengths in a predefined area, using RFView\textsuperscript{\tiny\textregistered}, a high-fidelity, site-specific, RF modeling \& simulation tool. We produce heatmap tensors from the radar returns, in range, azimuth [and Doppler], of the normalized adaptive matched filter (NAMF) test statistic. We then train a regression convolutional neural network (CNN) to estimate target locations from these heatmap tensors, and we compare the target localization accuracy of this approach with that of peak-finding and local search methods. This empirical study shows that our regression CNN achieves a considerable improvement in target location estimation accuracy. The regression CNN offers significant gains and reasonable accuracy even at signal-to-clutter-plus-noise ratio (SCNR) regimes that are close to the breakdown threshold SCNR of the NAMF. We also study the robustness of our trained CNN to mismatches in the radar data, where the CNN is tested on heatmap tensors collected from areas that it was not trained on. We show that our CNN can be made robust to mismatches in the radar data through few-shot learning, using a relatively small number of new training samples.
\end{abstract}

\end{spacing}
\vfill
\singlespacing % Reset line spacing to 1 from here on
\noindent \footnotesize Data availability: The data supporting the findings of this study was obtained using RFView\textsuperscript{\tiny\textregistered}. Conflict of Interest: none declared. \\
Authors' addresses: Shyam Venkatasubramanian and Vahid Tarokh are with the Department of Electrical and Computer Engineering, Duke University, Durham, NC 27705, USA, E-mail: (shyam.venkatasubramanian@duke.edu; vahid.tarokh@duke.edu); Sandeep Gogineni is with Information Systems Laboratories Inc., Dayton, OH 45431, USA (sgogineni@islinc.com); Bosung Kang is with the University of Dayton Research Institute, Dayton, OH 45469, USA (bosung.kang@udri.udayton.edu); Ali Pezeshki is with the Department of Electrical and Computer Engineering, Colorado State University, Fort Collins, CO 80523, USA, E-mail: (ali.pezeshki@colostate.edu); Muralidhar Rangaswamy is with the United States Air Force Research Laboratory, Wright-Patterson Air Force Base, OH 45433, USA, E-mail: (muralidhar.rangaswamy@us.af.mil).

\begin{spacing}{1.3} \normalsize

\section{\textbf{INTRODUCTION}}
Target localization is a crucial aspect in the design of modern radar systems, being foundational to radar applications such as surveillance, navigation, and military operations. Target localization is often performed post adaptive radar detection, where received radar signals are coherently processed to first detect the target and then estimate its location through model fitting. However, adaptive radar processing methodologies have several limitations, including sensitivity to errors in estimating the clutter covariance matrix \cite{kraut_adaptive,melvin_stap96} and high computational cost \cite{guerci2003space,guerci_stap00}. 

Expanding on the challenge of estimating the clutter covariance matrix, this crucial parameter is traditionally estimated using a limited number of returns from a radar dwell. The assumption that the clutter scattering phenomenon is a wide-sense stationary process --- which is valid only over short dwell times --- constrains the number of radar returns that can be used for covariance matrix estimation \cite{melvin_stap96}. This results in a mismatch between the true and estimated clutter statistics, which degrades both target detection and target localization performance \cite{raghavan_CFAR}. This issue is greatly exacerbated as the dimensionality of the radar array increases.

The availability of satellite topographic maps has made it possible to develop simulators that synthesize high-fidelity radar data from different regions. These simulators in turn alleviate the data paucity problem and enable the use of model-free, data-driven approaches (including deep learning), which often require large volumes of training data, for radar detection and estimation. One such simulator is RFView\textsuperscript{\tiny\textregistered} --- a ray tracing simulator that works based on a splatter, clutter, and target signal (SCATS) phenomenology engine, and offers many degrees of freedom on both transmit and receive to synthesize radar data \cite{gogineni_RFView}. We use the data generated by this simulator in this paper. Such simulators may eventually be used to generate large benchmarking datasets for radar, akin to image datasets such as CIFAR-10/100 \cite{krizhevsky2009learning} and ImageNet \cite{ILSVRC15} in computer vision, which have accelerated the development and evaluation of machine learning algorithms.

It has been well established that the performance of covariance-based radar estimation methods deteriorates rapidly below a ``breakdown threshold SCNR''. This threshold can be predicted in the asymptotic regime with random matrix theory \cite{Benaych_eigenvalues,Gogineni_Passive,Nadakuditi_correlation}. An important goal of this paper is to investigate empirically whether classical covariance-based methods can be augmented with deep neural networks to provide improved target localization post adaptive radar detection, especially in scenarios where classical covariance-based methods yield poor performance. To this end, we use RFView\textsuperscript{\tiny\textregistered} to generate a large set of radar returns for different target locations and strengths in the presence of high-fidelity clutter returns, within a predefined radar processing region. 

The test statistic we consider in this analysis is the generalized likelihood ratio (GLR) for a normalized adaptive matched filter (NAMF) \cite{rangaswamy_NAMF,conteNAMF}, which is widely used in radar detection. The NAMF test statistic coincides with the adaptive coherence estimator (ACE) \cite{kraut_adaptive, scharf_adaptive, scharf1996adaptive, kraut1997canonical}, when the clutter and noise distributions are Gaussian. 

Building upon this test statistic, we construct a multi-dimensional array of tests statistic values across range, azimuth, and Doppler, producing a ``heatmap tensor'' of test statistic values. Our objective is to design a deep neural network --- a regression convolutional neural network (CNN) --- that can accurately estimate the location (and the velocity) of radar targets from these heatmap tensors. We investigate the sensitivity of our regression CNN to mismatches in the radar data by training the CNN on the heatmap tensors from one location and testing it on the heatmap tensors from nearby locations (we generate radar data corresponding to several displaced radar scenarios). We also propose a few-shot learning approach to improve the robustness of our regression CNN. For context, and with the aforementioned heatmap tensors, we assess the asymptotic performance and the breakdown threshold SCNR of the NAMF test statistic, below which the performance of the NAMF rapidly degrades. We show that the CNN achieves improved localization accuracies relative to peak-finding and local search methods for SCNR below this breakdown threshold.

\emph{\textbf{Main contributions:}} The main contributions of this paper are as follows.
\begin{enumerate}[label=\textbf{\arabic*})]
    \item We empirically determine the breakdown threshold SCNR of the NAMF test statistic for target localization in the finite sample regime, and compare it with the asymptotic predictions of the breakdown threshold from random matrix theory.
    \item We illustrate that augmenting classical adaptive radar processing with CNNs can considerably improve target localization accuracy in the absence of mismatches in the radar data.
    \item We investigate the impacts of radar data mismatches, caused by displacing radar scenarios, on the CNN's target localization accuracy. We show that as the chordal distance between the clutter subspaces of displaced scenarios increases, the CNN's localization accuracy declines.
    \item We employ few-shot learning to make the CNN robust to data mismatches using a relatively small number of new data samples, without the need to extensively retrain the CNN.
\end{enumerate}

\emph{\textbf{Organization:}} The paper is organized as follows. 
In Section \ref{Sec3}, we provide a brief review of RFView\textsuperscript{\tiny\textregistered} and describe our matched, mismatched, and Doppler case radar scenarios that are examined in the paper. In Section \ref{Sec4}, we describe the generation of heatmap tensors for training and testing our regression CNNs for target localization. In Section \ref{Sec6}, we present our regression CNNs for target localization. In Section \ref{Sec5}, we empirically determine the breakdown threshold SCNR of the NAMF test statistic and compare it with the asymptotic theoretical prediction. In Section \ref{Sec7}, we provide numerical results for the matched case, and demonstrate the improvement in target localization accuracy afforded by our proposed regression CNN. In Section \ref{Sec8}, we provide numerical results for the mismatched case, and show that as the chordal distance between the clutter subspaces of displaced radar scenarios increases, the localization accuracy of the CNN decreases. Subsequently, we illustrate how few-shot learning can be used to improve network generalization in the mismatched case. In Section \ref{Sec9}, we provide additional numerical results in which Doppler processing is considered. Finally, in Section \ref{Sec10}, we present conclusions.

This paper is an expanded version of our prior 2022 IEEE Radar Conference paper \cite{Shyam_STAP}. The agreement between the finite sample performance of the NAMF test statistic and the performance guarantees of random matrix theory in the threshold SCNR region is provided in the current paper but not in \cite{Shyam_STAP}, which solely considers the performance of the MVDR beamformer \cite{capon_mvdr}. Furthermore, the current paper contains a more extensive performance evaluation of our CNN framework than what was included in the conference paper. The journal paper also includes a more extensive analysis of mismatched scenarios.

\section{\textbf{SIMULATION SCENARIOS}} \label{Sec3}

In this section, we describe the example radar scenarios that we have generated using  RFView\textsuperscript{\tiny\textregistered} for performance evaluation and benchmarking.

\subsection{\textbf{RFView\texorpdfstring{\textsuperscript{\tiny\textregistered}}{®} Modeling and Simulation Tool}} \label{Sec3.1}
RFView\textsuperscript{\tiny\textregistered} \cite{gogineni_RFView}, developed by ISL Inc, is a knowledge-aided, site-specific, physics-based, RF modeling and simulation platform that utilizes a world-wide database of terrain and land cover data for RF ray tracing simulation. Using this simulation environment, one can generate radar datasets for use in various signal processing applications. RFView\textsuperscript{\tiny\textregistered} has been validated against measured datasets spanning from VHF to K\textsubscript{a} band, with one specific example presented in \cite{gogineni_RFView}. 

\begin{table}[h!]
\caption{Shared Site and Radar Parameters}
\label{radar parameters}
\centering
\fontsize{10.7}{13}\selectfont
\begin{tabularx}{\columnwidth}{X|X}
\hline Parameters & Values \\ \hline
Carrier frequency $(f_\text{c})$ & $10,000 \ \text{MHz}$ \\
Bandwidth $(B)$ \& PRF $(f_p)$ & $5 \ \text{MHz}$ \& $1100 \ \text{Hz}$  \\
Transmitting \& Receiving antenna & $48 \times 5$ (horizontal $\times$ vertical elements) \\
Antenna element spacing & $0.015 \ \text{m}$ \\
Platform height & $1000 \ \text{m}$ \\
Area latitude (min, max) & $(32.4611^{\circ},32.6399^{\circ})$ \\
Area longitude (min, max) & $(-117.1554^{\circ},-116.9433^{\circ})$
\end{tabularx}
\end{table}

We consider three simulation scenarios, in which an airborne radar is looking over a coastal region in Southern California. The simulation region covers a $20 \ \text{km} \times 20 \ \text{km}$ area. RFView\textsuperscript{\tiny\textregistered} aggregates the information on land types and the geographical characteristics across the simulation region. The radar operates in `spotlight' mode and points toward the center of the simulation region. The values of the radar system parameters are listed in Table \ref{radar parameters}. We take the radar platform to be stationary during the data generation period for simplicity. In the first two examples, we do not consider Doppler processing and the radar data is generated from one return. The third example involves Doppler processing from multiple returns. The radar return is beamformed for each size $(48/L \times 5)$ receiver sub-array, which condenses the receiver array to size $(L \times 1)$.

\subsection{\textbf{Matched Case RFView\texorpdfstring{\textsuperscript{\tiny\textregistered}}{®} Example Scenario}} \label{Sec3.2}

\begin{figure}[h!]
    \centering
    \captionsetup{justification=centering}
    \includegraphics[width=0.88\linewidth]{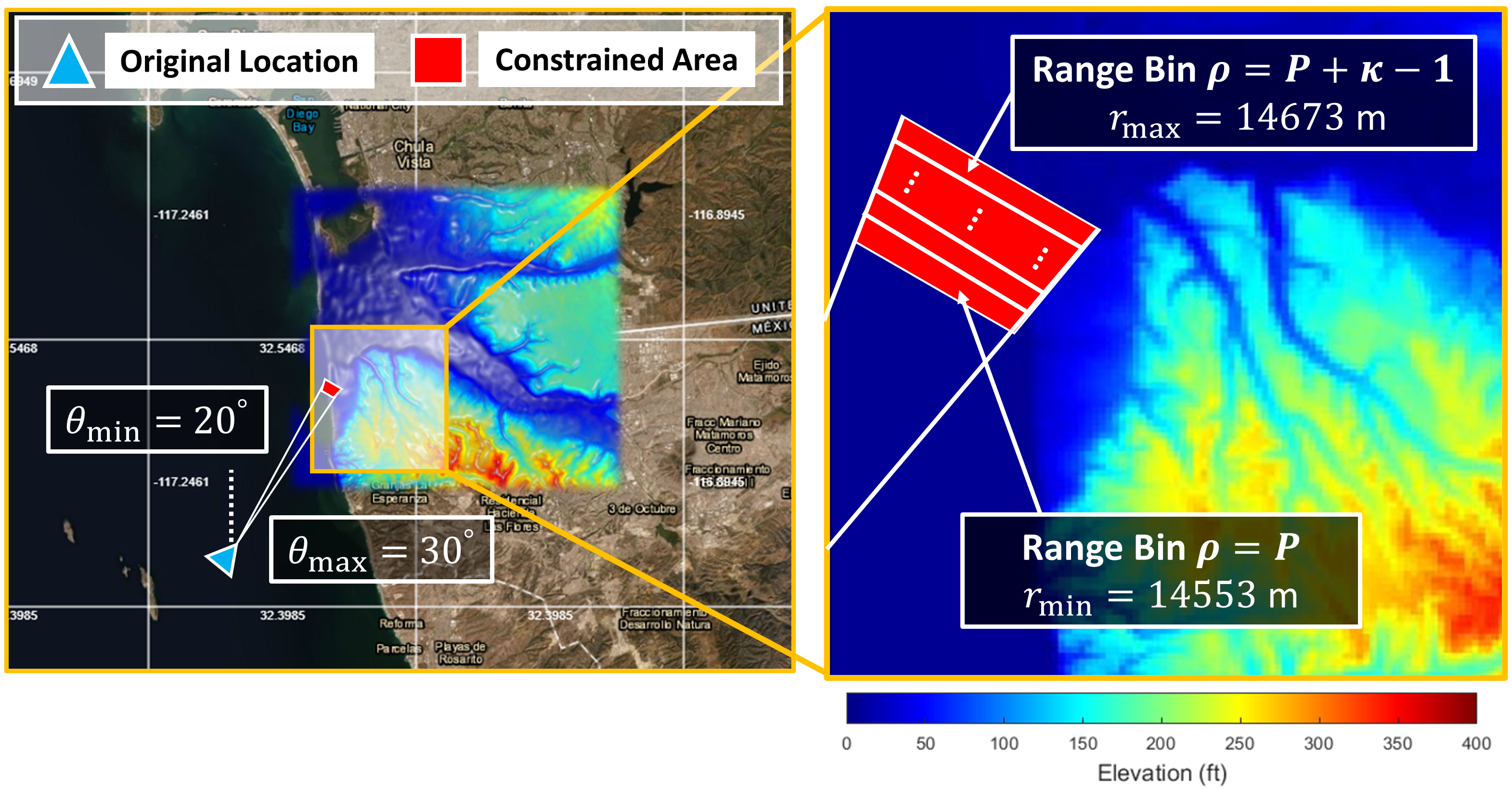}
    \caption{The map of the matched case RFView\textsuperscript{\tiny\textregistered} example scenario. The blue triangle is the platform location and the red region is the range-azimuth area for radar processing. The elevation heatmap overlaying the left image depicts the simulation region.}
    \label{matched map}
\end{figure}

Consider the red trapezoidal area in Figure~\ref{matched map}, which consists of $\kappa$ range bins and denotes the ground scene. The minimum and maximum range of this area from the radar platform and the azimuth angle bounds of this area are shown in Table~\ref{matched parameters}, where $r_{min}$ and $r_{max}$ denote the midpoints of range bins $P$ and $P + \kappa - 1$, respectively. The size of each range bin is $\Delta r = \frac{c}{2B} = 30 \ [m]$, where $c$ is the speed of light and $B$ is the bandwidth of the radar waveform. Range bin $\rho \in \{P,P+1,...,P+\kappa - 1\}$ is defined for $r \in [r_\rho-\Delta r/2,r_\rho+\Delta r/2] = [G + \rho \Delta r,G + (\rho+1) \Delta r]$, where $r_\rho$ is the midpoint of range bin $\rho$, and $G$ is the distance between the platform location and the simulation region. We run a set of $N$ independent experiments, where in each experiment, a point target is placed in range and azimuth, uniformly at random, inside the red ``radar processing'' region. We select the radar cross section (RCS)  of each target independently at random from the uniform distribution, $\mathcal{U}[\mu-l/2,\mu-l/2]$, where $\mu$ denotes the mean target RCS (dBsm). 

For each target placement, we generate $K$ independent random realizations of the radar return. We specify the values of the above parameters in Section \ref{Sec7}.

\begin{table}[h!]
\caption{Matched Case Site and Radar Parameters}
\label{matched parameters}
\centering
\fontsize{10.7}{13}\selectfont
\begin{tabularx}{\columnwidth}{X|X}
\hline \textbf{Original Location} -- Parameters & Values \\ \hline
Platform latitude, longitude & $32.4005^{\circ},-117.1993^{\circ}$\\
Constrained area range $(r_{\text{min}},r_{\text{max}})$ & $(14553 \ \text{m},14673 \ \text{m})$ \\
Constrained area azimuth $(\theta_{\text{min}},\theta_{\text{max}})$ & $(20^{\circ},30^{\circ})$ \\
\end{tabularx}
\end{table}

\subsection{\textbf{Mismatched Case RFView\texorpdfstring{\textsuperscript{\tiny\textregistered}}{®} Example Scenarios}} \label{Sec3.3}

\begin{figure}[h!]
    \centering
    \captionsetup{justification=centering}
    \includegraphics[width=0.8\linewidth]{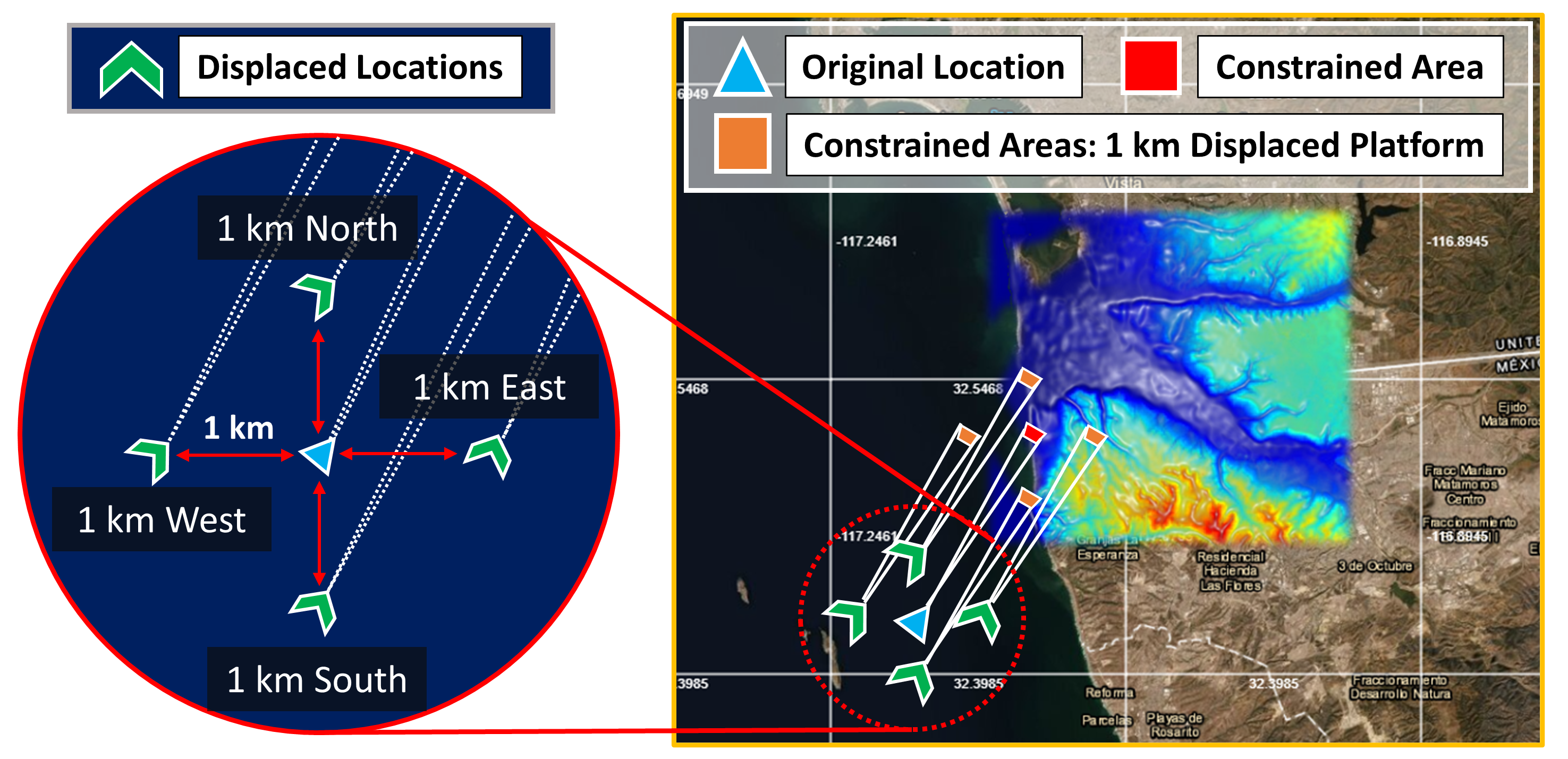}
    \caption{The map of the mismatched case RFView\textsuperscript{\tiny\textregistered} example scenario. Each of the displaced platform locations (1 km North, 1 km South, 1 km East, and 1 km West) are shown in green, and their relative range-azimuth areas for radar processing are depicted in orange. The original platform location (blue) and range-azimuth area (red) are recycled from the matched case.}
    \label{mismatched map}
\end{figure}

To study sensitivity to mismatch between the training and test environment, we consider several scenarios in which the radar platform is displaced relative to the matched case example scenario. Accordingly, the airborne radar platform looks at different range-azimuth areas when compared to the prior example. The prior example will serve as the baseline (or original) scenario, and we denote this scenario by \textbf{(O)}. We consider $1$ [km] displacements of the airborne platform relative to its location in \textbf{(O)} in each cardinal direction: North, West, South, and East. Correspondingly, we denote these scenarios by \textbf{(N)}, \textbf{(W)}, \textbf{(S)}, and \textbf{(E)}, respectively. The platform locations for these four `mismatched' scenarios are displayed in Figure \ref{mismatched map} and their respective range-azimuth areas (the orange trapezoidal regions) for radar processing are listed in Table~\ref{mismatched parameters}. We consider $0.1N$ independent experiments to generate test data for the mismatched cases. In each experiment, we place a point target uniformly at random inside the corresponding orange region. We select the radar cross section (RCS)  of each target independently at random from the uniform distribution, $\mathcal{U}[\mu-l/2,\mu-l/2]$. We generate $K$ independent random realizations of the radar return for each target placement, and we specify the values of the above parameters in Section \ref{Sec8}.

\begin{table}[h!]
\caption{Mismatched Case Site and Radar Parameters}
\label{mismatched parameters}
\centering
\fontsize{10.7}{13}\selectfont
\begin{tabularx}{\columnwidth}{X|X}
\hline \textbf{1 km North (N)} -- Parameters & Values \\ \hline
Platform latitude, longitude & $32.4095^{\circ},-117.1993^{\circ}$\\
Constrained area range $(r_{\text{min}},r_{\text{max}})$ & $(13800 \ \text{m},13920 \ \text{m})$ \\
Constrained area azimuth $(\theta_{\text{min}},\theta_{\text{max}})$ & $(20^{\circ},30^{\circ})$ \\
\hline \textbf{1 km West (W)} -- Parameters & Values \\ \hline
Platform latitude, longitude & $32.4005^{\circ},-117.2099^{\circ}$\\
Constrained area range $(r_{\text{min}},r_{\text{max}})$ & $(15207 \ \text{m},15327 \ \text{m})$ \\
Constrained area azimuth $(\theta_{\text{min}},\theta_{\text{max}})$ & $(20^{\circ},30^{\circ})$ \\
\hline \textbf{1 km South (S)} -- Parameters & Values \\ \hline
Platform latitude, longitude & $32.3915^{\circ},-117.1993^{\circ}$\\
Constrained area range $(r_{\text{min}},r_{\text{max}})$ & $(15321 \ \text{m},15441 \ \text{m})$ \\
Constrained area azimuth $(\theta_{\text{min}},\theta_{\text{max}})$ & $(20^{\circ},30^{\circ})$ \\
\hline \textbf{1 km East (E)} -- Parameters & Values \\ \hline
Platform latitude, longitude & $32.4005^{\circ},-117.1887^{\circ}$\\
Constrained area range $(r_{\text{min}},r_{\text{max}})$ & $(13921 \ \text{m},14041 \ \text{m})$ \\
Constrained area azimuth $(\theta_{\text{min}},\theta_{\text{max}})$ & $(20^{\circ},30^{\circ})$ \\
\end{tabularx}
\end{table}

\subsection{\textbf{Doppler Processing RFView\texorpdfstring{\textsuperscript{\tiny\textregistered}}{®} Example Scenario}} \label{Sec3.4}
For our Doppler processing example scenario, we consider a scenario similar to the matched case example scenario outlined in Section \ref{Sec3.2}. However, each target now has a relative velocity (relative to the radar platform). We select the velocities uniformly at random from the interval $[v_{\text{min}},v_{\text{max}}] = [175 , 190]$ [m/s]. The dataset generation procedure follows the procedure outlined in Section \ref{Sec3.2}, and we specify the values of all relevant parameters in Section \ref{Sec9}.

\section{\textbf{HEATMAP TENSOR GENERATION}} \label{Sec4}

In this section, we describe the generation of heatmap tensors using the NAMF test statistic. These heatmap tensors comprise the input data samples for our regression CNNs.

\subsection{\textbf{NAMF Test Statistic}} \label{Sec4.1}
Suppose $X_\rho, Z_\rho \in \mathbb{C}^{\Lambda L}$ are the random variables describing the target and clutter-plus-noise information, and suppose that $x^k_\rho \sim X_\rho$ and $z^k_\rho \sim Z_\rho$ denote the target and clutter-plus-noise data for the $k^{\text{th}}$ independent random realization of the radar return. Recording $x^k_\rho, z^k_\rho$, $\forall k \in \{1,\ldots,K\}$, we obtain $\mathbf{X}_\rho, \mathbf{Z}_\rho \in \mathbb{C}^{\Lambda L \times K}$. We let $\mathbf{Y_\rho}\in \bbC^{\Lambda L \times K}$ be the matched filtered radar return data matrix associated with range cell (bin) $\rho$. The columns of $\mathbf{Y_\rho}$ are $\Lambda L \times 1$ dimensional vectors that are built by stacking $L$-dimensional array measurements, obtained using $\Lambda$ pulse transmissions. For each pulse return, we have $K$ independent realizations. The radar return data matrix is given by: 
\begin{equation}
\mathbf{Y_\rho}=\beta_{\rho}\mathbf{X_\rho} + \mathbf{Z_\rho}\label{eq_alternative}
\end{equation}
Where $\beta_\rho=1$ if range bin $\rho$ contains a target and $\beta_\rho=0$ otherwise, $\mathbf{X}_\rho$ is the target response matrix, and $\mathbf{Z_\rho}$ is the clutter-plus-noise response matrix. 

Let $\mathbf{a_\rho}(\theta, \phi)\in \mathbb{C}^L$ denote the receiver array steering vector, where $\theta$ is the azimuth angle variable and $\phi$ is the elevation angle variable. Let $\mathbf{d_\rho}(v)$ be the Doppler processing linear phase vector across $\Lambda$ pulses, with $\mathbf{d}_\rho(v) = \left[ 1, e^{-i 2 \pi f_d/f_p}, \ldots, e^{-i 2 \pi  (\Lambda - 1)f_{d}/f_p} \right]^T$, where $f_{\text{d}}$ is the Doppler shift frequency and $f_p$ is the pulse repetition frequency (PRF). The effective space-time steering vector, $\mathbf{\tilde{a}_\rho}(\theta,\phi,v)$, in azimuth, elevation, and Doppler for radar processing is:
\begin{equation}
\mathbf{\tilde{a}}_\rho (\theta, \phi, v) =  \mathbf{d}_\rho(v) \otimes \mathbf{a}_\rho(\theta, \phi)
\end{equation}
Where $\otimes$ denotes Kronecker product. Let $\mathbf{\Sigma_\rho} = \mathbb{E}[Z_\rho Z_\rho^H] \in \mathbb{C}^{(\Lambda L) \times (\Lambda L)}$ denote the true clutter-plus-noise covariance matrix determined from the random variable $Z_\rho$. Consequently, the sample clutter-plus-noise covariance matrix, $\mathbf{\hat{\Sigma}_\rho}$, is calculated as: $\mathbf{\hat{\Sigma}_\rho} = (\mathbf{Z_\rho} \mathbf{Z_\rho}^H) / K$. The NAMF test statistic, $\Gamma_\rho(\theta, \phi,v)$, at coordinates $(\theta, \phi, v)$ in range bin $\rho$, is given by \cite{rangaswamy_NAMF, conteNAMF}:
\begin{align}
    \Gamma_\rho(\theta,\phi,v)= \frac{\| \mathbf{\Tilde{a}_\rho}(\theta, \phi, v)^H \mathbf{\hat{\Sigma}_\rho}{\vphantom{\Sigma}}^{-1} \mathbf{Y}_\rho\|_2^2}{\mathbf{\Tilde{a}_\rho}(\theta, \phi, v)^H \mathbf{\hat{\Sigma}_\rho}{\vphantom{\Sigma}}^{-1} \mathbf{\Tilde{a}_\rho}(\theta, \phi, v) \ \|\text{diag}(\mathbf{Y}_\rho^H \mathbf{\hat{\Sigma}_\rho}{\vphantom{\Sigma}}^{-1} \mathbf{Y}_\rho)\|_2
    }
\end{align}
When the airborne radar platform is at a long range from the ground scene, the elevation angle, $\phi$, is effectively zero. This is the case in radar scenarios that we consider in this paper. 

The NAMF test statistic coincides with the adaptive coherence (cosine) estimator (ACE) \cite{kraut_adaptive, scharf_adaptive, scharf1996adaptive, kraut1997canonical}, when the clutter and noise distributions are Gaussian. 

\subsection{\textbf{NAMF Heatmap Tensor Generation}} \label{Sec4.2}

\textbf{\emph{Without Doppler:}} For stationary target simulation scenarios (see Sections \ref{Sec3.2} and \ref{Sec3.3}), the Doppler frequency shift is zero, and accordingly, we only consider a single pulse transmission ($\Lambda=1$) and replace the effect of coherent processing across multiple pulses with a controlled signal-to-clutter-plus-noise ratio (SCNR). We set the elevation angle to $\phi=0$, because in each of our scenarios, the radar platform is far from the ground scene. We sweep the radar steering vector, $\mathbf{\tilde{a}}_\rho (\theta, 0, 0)$, across azimuth angle with step size $\Delta\theta$, from $\theta_{\text{min}}$ to $\theta_{\text{max}}$. This produces an NAMF azimuth spectrum at each range bin. Arranging these spectra across $\kappa$ range bins gives a size $\kappa\times (\lfloor{(\theta_{\text{max}} - \theta_{\text{min}})/\Delta\theta}\rfloor + 1)$ matrix of NAMF test statistics. 

\begin{figure}[h!]
    \centering
    \captionsetup{justification=centering}
    \includegraphics[width=0.9\linewidth]{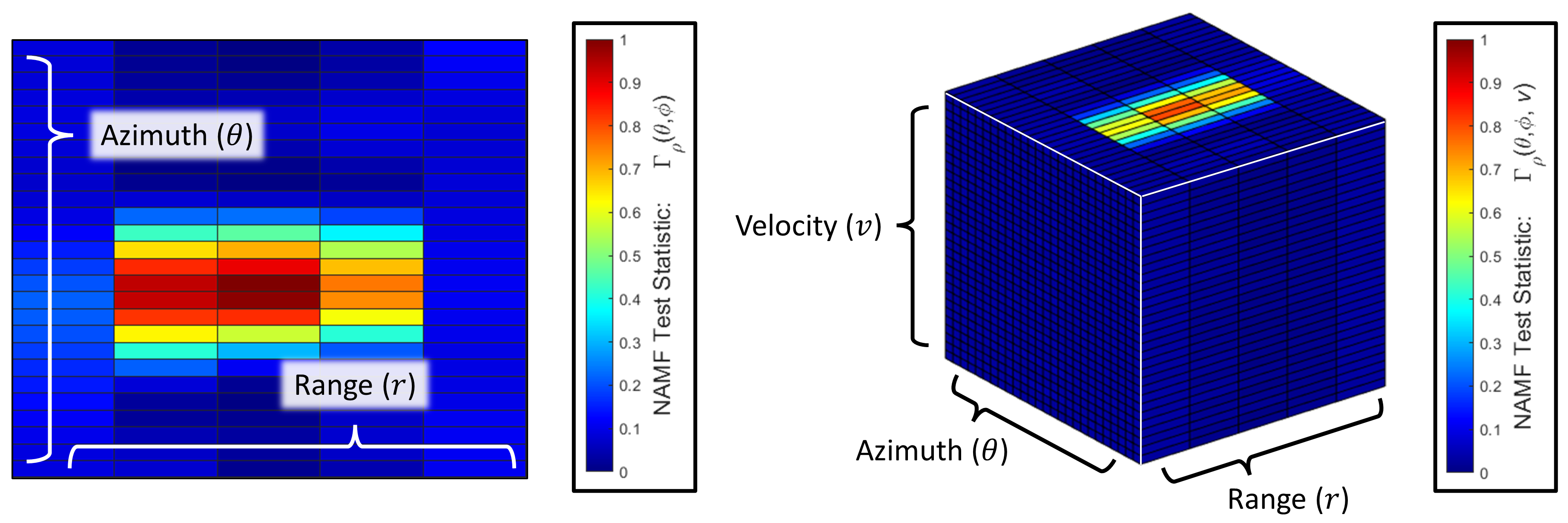}
    \caption{(Left) Size $5\times26$ heatmap tensor example from the matched case RFView\textsuperscript{\tiny\textregistered} example scenario (see Section \ref{Sec3.2}). Each size $1 \times 26$ `array in azimuth' pertains to one of the $\kappa = 5$ unique range bins. (Right) Size $5 \times 26 \times 31$ heatmap tensor example from the Doppler processing RFView\textsuperscript{\tiny\textregistered} example scenario (see Section \ref{Sec3.4}). Each size $26 \times 31$ heatmap image has an azimuth and velocity dimension, and pertains to one of the $\kappa = 5$ unique range bins.}
    \label{matched case heatmap}
\end{figure}

The values of $\theta_{\text{min}}$ to $\theta_{\text{max}}$ are given in Tables \ref{matched parameters} and \ref{mismatched parameters} for our various simulation scenarios. We build these heatmap matrices for $N$ independent random placements of a point target inside the radar processing area (red trapezoidal region in Figure \ref{matched map}). The collection of these $N$ matrices constitutes our dataset, which is divided into training and validation examples for our regression CNN. An example of a heatmap matrix is given in Figure \ref{matched case heatmap} (left), for $(\Delta r,\Delta \theta) = (30  \ \text{m},0.4^{\circ})$, $\kappa = 5$. The heatmap matrix in this case is of size $5 \times 26$. The target coordinates, relative to the radar platform, are  $(r^*,\theta^*) = (14606 \ \text{m}, 25.7494^{\circ})$.

\textbf{\emph{With Doppler:}} In the scenario where we consider Doppler (see Section \ref{Sec3.3}), we use $\Lambda=4$ pulses for Doppler processing. We sweep the radar steering vector, $\mathbf{\tilde{a}}_\rho (\theta, 0, v)$, across azimuth angle with step size $\Delta\theta$, from $\theta_{\text{min}}$ to $\theta_{\text{max}}$, and across velocity (relative to the radar platform), with step size $\Delta v$, from $v_{\text{min}}$ to $v_\text{max}$. This produces an NAMF azimuth-Doppler heatmap at each range bin. Arranging these heatmaps across $\kappa$ range bins gives a size $\kappa \times \lfloor{(\theta_{\text{max}} - \theta_{\text{min}})/\Delta\theta}\rfloor \times \lfloor{(v_{\text{max}} - v_{\text{min}})/\Delta v}\rfloor$ tensor of NAMF test statistics. 

We build these heatmap tensors for $N$ independent random placements of a point target inside the radar processing area (red trapezoidal region in Figure \ref{matched map}). The collection of these $N$ tensors constitutes our dataset for training and evaluating our regression CNN. An example of a heatmap tensor is given in Figure \ref{matched case heatmap} (right), for $(\Delta r,\Delta \theta, \Delta v) = (30  \ \text{m},0.4^{\circ},0.5 \ \text{m/s})$, $\kappa = 5$ range bins. In this case the heatmap tensor is of size $5 \times 26 \times 31$. The point target coordinates, relative to the radar platform,  in this example are $(r^*,\theta^*,v^*) = (14617 \ \text{m}, 24.5353^{\circ}, 181.5004 \ \text{m/s})$.

\section{\textbf{REGRESSION CNN ARCHITECTURES}} \label{Sec6}

In this section, we describe our regression CNN architectures for estimating the location and velocity (if Doppler processing is performed) of a single point target from heatmap matrix/tensor samples. The regressions CNNs are trained and validated with the NAMF heatmap matrix/tensor datasets described in Section \ref{Sec4.2}, where the total number of data samples we use for training and validation is $N = 10^5$. Of these, $N_{train}=0.9N$ points are used for training and $N_{val}=0.1N$ are used for validation. The CNN architectures are extensions of the LeNet-5 architecture \cite{lecun1998gradient}, adapted for our considered heatmap matrix/tensor sizes.

We now define the error metric to interpret the localization accuracies of our regression CNN framework and the more classical approaches. We consider a dataset of heatmap tensors produced using the NAMF. Let $(\thickbreve{r}_j, \thickbreve{\theta}_j, \thickbreve{v}_j)$ denote the range, azimuth, and velocity of the midpoint of the peak heatmap tensor cell, let $(\thickdot{\theta}_j)$ and $(\thickdot{v}_j)$ be the optimal azimuth and velocity values yielded by each local search algorithm, and let $(\thicktilde{r}_j, \thicktilde{\theta}_j, \thicktilde{v}_j)$ denote the azimuth and velocity values predicted by the regression CNN, for some example $j$ from our dataset. While our heatmap tensors follow the polar coordinate system, we can transform their ground truth target locations into Cartesian coordinates to determine the localization error in meters, whereby $(\thickbreve{r}_j, \thickbreve{\theta}_j, \thickbreve{v}_j) \rightarrow (\thickbreve{x}_j, \thickbreve{y}_j, \thickbreve{v}_j)$ and $(\thicktilde{r}_j, \thicktilde{\theta}_j, \thicktilde{v}_j) \rightarrow (\thicktilde{x}_j, \thicktilde{y}_j, \thicktilde{v}_j)$. Let $(r_j^*,\theta_j^*,v_j^*) \rightarrow (x_j^*,y_j^*,v_j^*)$ denote the ground truth target location and velocity for example $j$ from our dataset. For vector (or scalar) inputs $\mathbf{s}_j^*,\thickhat{\mathbf{s}}_j$, the localization (or attribute estimation) error, $Err(\mathbf{s}_j^*,\thickhat{\mathbf{s}}_j)$, over the $N_{val}$ validation examples, is defined as:
\begin{align}
    Err(\mathbf{s}_j^*,\thickhat{\mathbf{s}}_j) = \frac{\sum_{j = 1}^{N_{val}} \| \mathbf{s}_j^* - \thickhat{\mathbf{s}}_j\|_2}{N_{val}} \label{eq_CNN_loc}.
\end{align}
In the matched and mismatched cases, for the localization error, we let $(\mathbf{s}_j^*,\thickhat{\mathbf{s}}_j) = ([x_j^*,y_j^*], [\thickhat{x}_j, \thickhat{y}_j])$, where $(\thickhat{x}_j, \thickhat{y}_j, \thickhat{v}_j) = (\thickbreve{x}_j, \thickbreve{y}_j, \thickbreve{v}_j)$ in the peak cell midpoint approach $[Err_{\text{NAMF}}]$, and $(\thickhat{x}_j, \thickhat{y}_j, \thickhat{v}_j) = (\thicktilde{x}_j, \thicktilde{y}_j, \thicktilde{v}_j)$ in the regression CNN approach $[Err_{\text{CNN}}]$. For the azimuth estimation error, we let $(\mathbf{s}_j^*,\thickhat{\mathbf{s}}_j) = (\theta_j^*, \thickhat{\theta}_j)$, where $\thickhat{\theta}_j = \thickbreve{\theta}_j$ in the peak cell midpoint approach $[(Err_{\text{NAMF}})_\theta]$, $\thickhat{\theta}_j = \thickdot{\theta}_j$ in the local search approach $[(Err_{\text{LS}})_\theta]$, and $\thickhat{\theta}_j = \thicktilde{\theta}_j$ in the regression CNN approach $[(Err_{\text{CNN}})_\theta]$. In the Doppler processing case, we also consider the velocity estimation error, letting $(\mathbf{s}_j^*,\thickhat{\mathbf{s}}_j) = (v_j^*, \thickhat{v}_j)$, where $\thickhat{v}_j = \thickbreve{v}_j$ in the peak cell midpoint approach $[(Err_{\text{NAMF}})_v]$, with $\thickhat{v}_j = \thickdot{v}_j$ in the local search approach $[(Err_{\text{LS}})_v]$, and $\thickhat{v}_j = \thicktilde{v}_j$ in the regression CNN approach $[(Err_{\text{CNN}})_v]$.

\subsection{\textbf{Regression CNN for Scenarios Without Doppler Processing}}

\begin{figure}[h!]
    \centering
    \captionsetup{justification=centering}
    \includegraphics[width=0.84\linewidth]{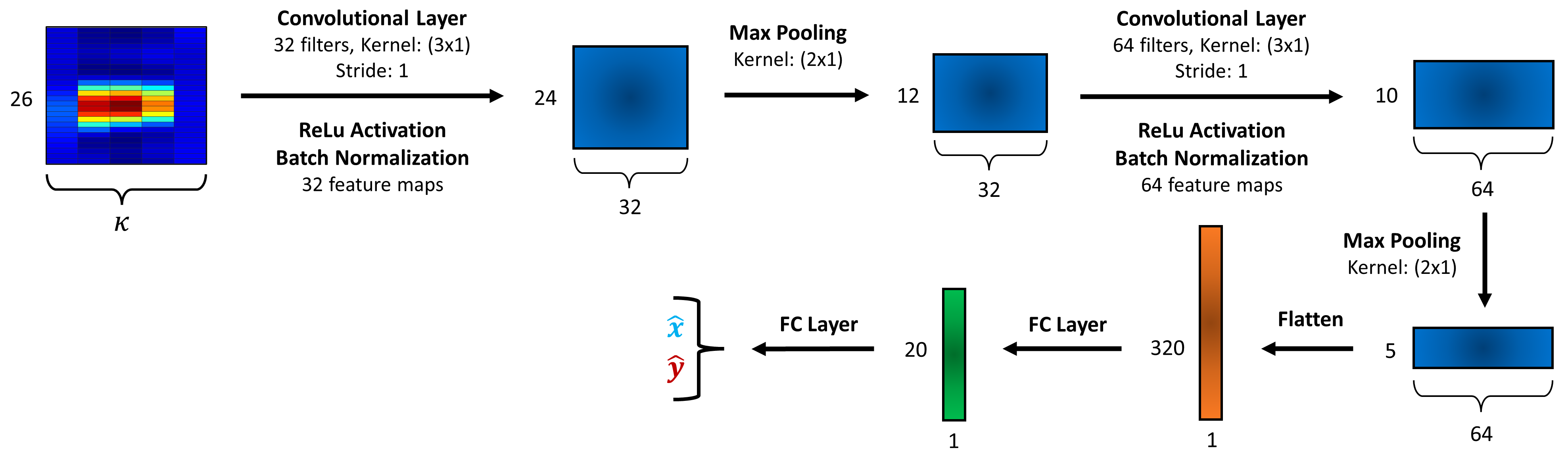}
    \caption{Baseline regression CNN architecture for azimuth step size $\Delta \theta = 0.4^{\circ}$ and depth parameter $\kappa$.}
    \label{CNN_default_parameters}
\end{figure}

The architecture of the regression CNN for the case that we do not do any Doppler processing is shown in Figure \ref{CNN_default_parameters}. In all of our examples, the azimuth extent of the radar processing areas, (i.e., $\theta_\text{max}-\theta_{\text{min}}$) is $10^{\circ}$, where the exact values of $\theta_\text{min}$ and $\theta_\text{max}$ are given in Tables \ref{matched parameters} and \ref{mismatched parameters}. We consider a grid with step size $\Delta\theta=0.4^{\circ}$ over the azimuth dimension. Our radar processing areas (the red and orange trapezoids in Figures \ref{matched map} and \ref{mismatched map}) consist of $\kappa=5$ range bins. The size of each range bin is equal to $\Delta r = 30 \ [m]$ and is selected to match the classical radar range resolution, determined by the bandwidth of the transmitted waveform. Consequently, each input data sample to the regression CNN is a $5 \times 26$ heatmap matrix.

Each input data sample first passes through a convolutional layer with kernel size $3 \times 1$ and stride 1, producing 32 feature maps, to which we apply ReLu activation and batch normalization. We apply max pooling to the output with kernel size $2 \times 1$. Next, we pass this output through another convolutional layer with kernel size $3 \times 1$ and stride 1, yielding 64 feature maps, to which we once again apply ReLu activation and batch normalization. After applying max pooling to this output with kernel size $2 \times 1$, we flatten and pass it through two fully connected layers to obtain an estimate $(\thickhat{r},\thickhat{\theta})$ of the target location in range-azimuth coordinates. We subsequently convert this estimate to an estimate $(\thickhat{x},\thickhat{y})$ in the Cartesian coordinates. The network in total has $13{\small,}374$ trainable parameters. We use the empirical average Euclidean distance as the loss function for training the regression CNN. The validation performance of the network is evaluated using the same metric, but over the $N_{val}$ validation samples.

% \ap{We use the empirical average Euclidean distance between the true target coordinates $\mathbf{s}_j^*$ in the $j$th sample and its corresponding estimate $\thickhat{\mathbf{s}}_j$, i.e.,
% \begin{align}
%     Err(\mathbf{s}_j^*,\thickhat{\mathbf{s}}_j) = \frac{\sum_{j = 1}^{N_{train}} \| \mathbf{s}_j^* - \thickhat{\mathbf{s}}_j\|_2}{N_{train}} \label{eq_CNN_loc}
%  \end{align}
% as the loss function for training the regression CNN. The validation performance of the network is evaluated using the same metric, but over the validation set, which has $N_{val}$ samples.}

\subsection{\textbf{Regression CNN for Scenarios With Doppler Processing}}

\begin{figure}[h!]
    \centering
    \captionsetup{justification=centering}
    \includegraphics[width=0.85\linewidth]{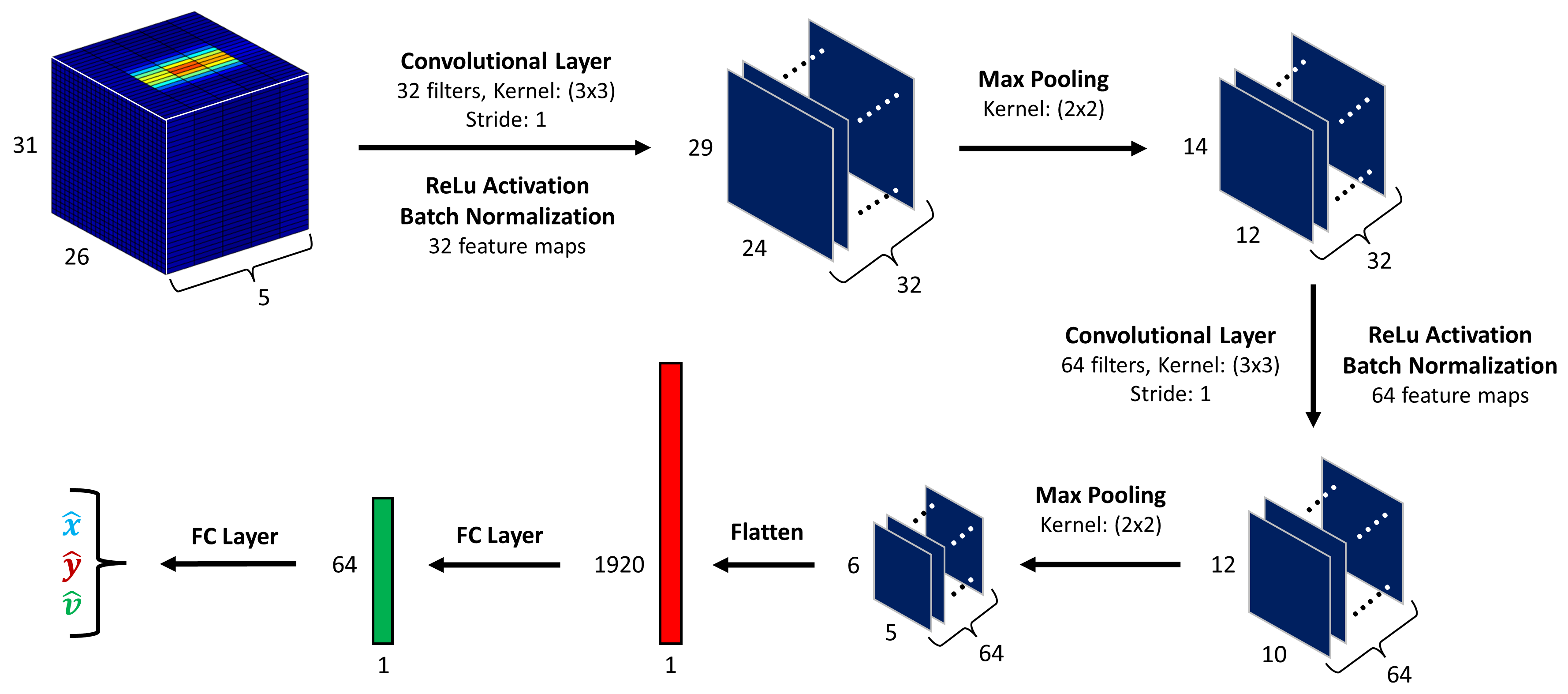}
    \caption{Doppler CNN architecture for default grid step size $(\Delta r, \Delta \theta, \Delta v) = (30 \ \text{m}, 0.4^{\circ}, 0.5 \ \text{m/s})$.}
    \label{CNN_Doppler}
\end{figure}

The structure of our regression CNN when Doppler processing is involved is shown in Figure \ref{CNN_Doppler}. The main difference is that the input data samples (NAMF heatmap matrices) are now 3D cubes in range, azimuth, and velocity. The range and azimuth extents and grid step sizes are similar to the previous case. We consider a grid step size of $\Delta v=0.5 \ \text{m/s}$, with $v_{\text{max}}-v_{\text{min}}=15 \ [m/s]$ in our examples. This results in heatmap tensors of size $5 \times 26 \times 31$ for the CNN. The number of layers and size of the regression CNN are selected based on the benchmark provided by LeNet-5 \cite{lecun1998gradient} for our specified input size. The network has $143{\small,}299$ trainable parameters. The processing steps in the network and the loss function are similar to that in the previous case.

\section{\textbf{EMPIRICAL RESULTS FOR MATCHED CASE}} \label{Sec7}

Our aim in this section is to examine the extent to which target localization can be improved by post-processing our heatmap matrices consisting of NAMF test statistics with deep learning, particularly near the NAMF breakdown threshold SCNR, where the performance of the NAMF test statistic (CNN input data samples) degrades rapidly. We start by determining the breakdown threshold SCNR of the NAMF and then move to the performance evaluation of our CNN. We do not consider any radar scenario mismatch in this section; that is, we evaluate the performance of the regression CNN on the same radar scenario on which it is trained, although using a separate validation dataset. The analysis of mismatched scenarios is considered in Section \ref{Sec8}.

% As outlined in Section \ref{Sec6}, in the matched case, we consider four evaluations of our regression CNN framework to estimate the position of a single target in the presence of clutter and noise using heatmap tensors produced with the NAMF test statistic. We note that each of the evaluations considered is scenario-specific and data-specific.

\subsection{\textbf{NAMF Breakdown Threshold SCNR}} \label{Sec5}

%Recent advances in random matrix theory (RMT) over the past decade have shown that in the asymptotic regime, 

It has been well established that covariance-based and subspace-based detection and estimation methods suffer sudden and severe performance degradation when the signal-to-noise-ratio (SNR) [or signal-to-clutter-plus-noise-ratio (SCNR)] drops below a given threshold. This phase transition in performance is caused by a swapping of the signal and noise subspaces in the SVD of the data matrix or its corresponding Gram matrix; that is, the dominant singular vector spaces no longer represent the correct subspace. The threshold at which this occurs with high probability is called the breakdown threshold. The threshold values depend on the specifics of the covariance-based or subspace-based detection and estimation method. The reader is referred to \cite{tufts1991threshold, thomas1995probability, pakrooh2016threshold} for calculations of breakdown thresholds for various estimation and detection methods. 

% perturbation models experience a severe degradation in performance below a threshold signal-to-noise ratio (SNR) --- referred to as the `phase transition threshold' --- with high probability \cite{Benaych_eigenvalues,Benaych_singular,Nadakuditi_correlation,Nadler_PCA}. 

In \cite{Gogineni_Passive}, Gogineni et al. provide an asymptotic analysis of the breakdown threshold SNRs of generalized likelihood ratio tests in Gauss-Gauss detection problems using random matrix theory. The authors show that as the size dimensions, $M$ and $N$, of the data matrix both go to infinity at the same rate (i.e. $M/N=c$, where $c$ is a constant), the breakdown threshold SNR approaches $c$ with high probability. Our data model from Eq.~\eqref{eq_alternative}, after whitening with the clutter-plus-noise covariance matrix, is consistent with the model considered in \cite{Gogineni_Passive}. The result of \cite{Gogineni_Passive} gives us an asymptotic prediction of the breakdown threshold SCNR, $c$, for the NAMF test statistic:
\begin{equation}
    c = 10\log_{10}\sqrt{\frac{(\Lambda \boldsymbol{\cdot} L)}{K}} \quad \text{(dB)}\label{eq_threshold}
\end{equation}

As we will now show, this asymptotic threshold value agrees reasonably well with the breakdown threshold SCNR of the NAMF test statistic in the finite sample regime, where the target localization error is computed using the midpoint of the peak NAMF heatmap matrix cell as the estimated target location. We first define the output SCNR for a point target at range bin $\rho$ as:
\begin{align}
    &(\text{SCNR}_{\text{Output}})_\rho = 10\log_{10} \left[ \frac{\Tr(\mathbf{{X}_\rho}{\vphantom{\Sigma}}^H \mathbf{\hat{\Sigma}_\rho}{\vphantom{\Sigma}}^{-1} \mathbf{{X}_\rho})}{\Tr(\mathbf{{Z}_\rho}{\vphantom{\Sigma}}^H \mathbf{\hat{\Sigma}_\rho}{\vphantom{\Sigma}}^{-1} \mathbf{{Z}_\rho})} \right]  \quad \text{(dB)} \label{eq_output_SCNR}
\end{align}
where $\mathbf{\hat{\Sigma}_\rho}$ is the sample clutter-plus-noise covariance matrix for range bin $\rho$. Averaging the output SCNR across the $N$ point target placements yields the mean output SCNR of our dataset. In this analysis, we vary the mean output SCNR from $-20$ dB to $20$ dB by changing the mean target RCS, $\mu$ (with $l = 10$ dBsm), for a fixed clutter-to-noise ratio (CNR) of $20$ dB. Furthermore, we let $\Lambda = 1, L = 16$, and $K \in \{100, 500\}$.

\begin{figure}[h!]
    \centering
    \captionsetup{justification=centering}
    \includegraphics[width=0.55\linewidth]{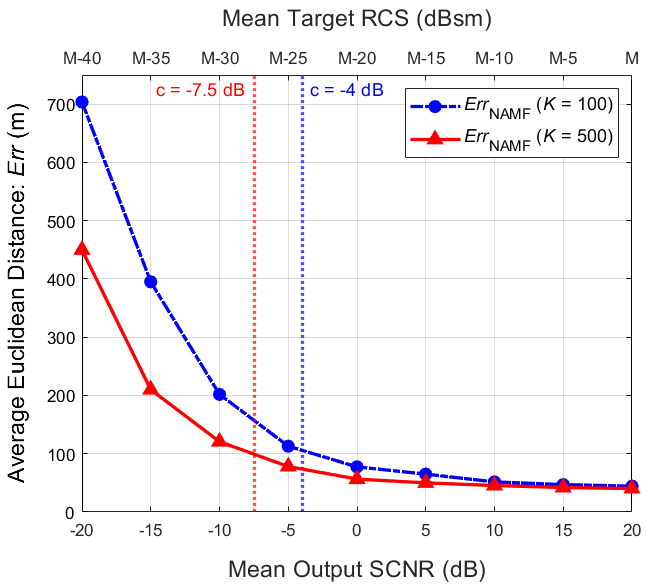}
    \caption{Threshold SCNR analysis of the NAMF test statistic for fixed clutter-to-noise ratio and variable mean target RCS. The mean target RCS, $\mu$, is selected so the mean output SCNR varies from $-20$ dB ($\mu = M-40$ dBsm) to $20$ dB ($\mu = M$ dBsm).}
    \label{Fixed_CNR_RCS_RMT}
\end{figure}

Figure \ref{Fixed_CNR_RCS_RMT} shows $Err_{\text{NAMF}}$, the mean $2$-norm of the error (see Section \ref{Sec6}) in estimating target locations using the peak cell midpoint approach, with respect to the mean output SCNR. The blue curve and the red curve, respectively, correspond to having $K=100$ and $K=500$ independent realizations of the radar return for each of the $N_{val}=9000$ random placements of a point target within the radar processing area. We observe in both plots that the localization performance of the NAMF test statistic rapidly degrades below the threshold SCNRs (the dotted vertical lines). The threshold values ($-4$ dB for $K=100$, $-7.5$ dB for $K=500$) are obtained from Eq. \eqref{eq_threshold}.

\subsection{\textbf{Evaluating Regression CNN for Variable Mean Output SCNR}} \label{Sec7.1}

In this section, we evaluate the target localization and azimuth estimation performance of our regression CNN across different mean output SCNR values. Let $\Lambda = 1, L = 16, K = 100$. We vary the mean output SCNR from $-20$ dB to $20$ dB by changing the mean target RCS, $\mu$ (with $l = 10$ dBsm), for a fixed CNR of $20$ dB. For each mean output SCNR value, we obtain a dataset consisting of $N = 9 \times 10^4$ heatmap matrix samples. We utilize $90\%$ of the dataset for training ($N_{train} = 8.1 \times 10^4$) and the remaining $10\%$ for validation ($N_{val} = 9 \times 10^3$). The entire dataset is from the same radar scenario, so there exists no mismatch between the training and validation datasets. We train the regression CNN from Figure \ref{CNN_default_parameters} on the training dataset using the Adam optimizer~\cite{kingma_14}, with learning rate hyperparameter $\alpha = 1 \times 10^{-3}$ (tuned via experimentation), and weight decay hyperparameter $\lambda = 1 \times 10^{-3}$ to prevent model overfitting ($\mathcal{L}_2$-regularization). 

\begin{figure*}[h!]
    \centering
    \captionsetup{justification=centering}
    \includegraphics[width=0.95\linewidth]{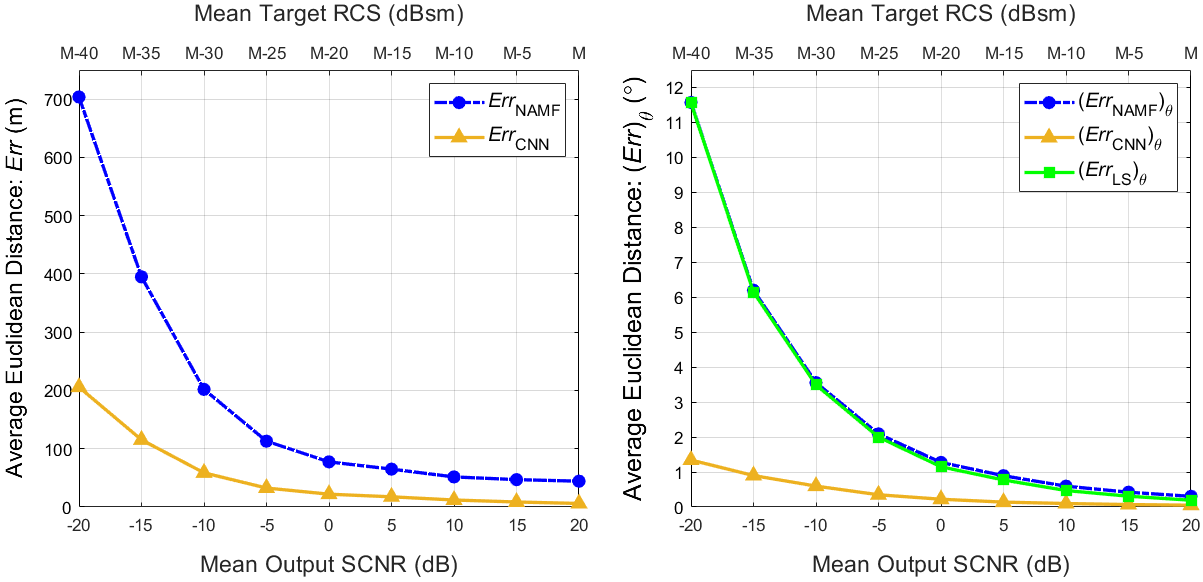}
    \caption{(Left) Target localization performance of the regression CNN (orange) vs SCNR and comparison with NAMF peak cell midpoint method (blue). (Right) Azimuth estimation accuracy of the regression CNN (orange) vs SCNR and comparison with azimuth estimation using NAMF peak cell midpoint (blue) and local search around the NAMF peak cell midpoint (green).}
    \label{Fixed_CNR_RMT_NAMF_Gain}
    \vspace{-1.5em}
\end{figure*}

Figure~\ref{Fixed_CNR_RMT_NAMF_Gain} (left) depicts the target localization accuracy of our regression CNN (in orange) at different mean output SCNR values, where the vertical axis marks the average Euclidean distance between the CNN predictions and the true target locations. The blue curve depicts the average Euclidean distance between the NAMF peak cell midpoint estimates and the true target locations for benchmarking performance gains. We recall that the radar range resolution, dictated by the waveform bandwidth, is $30$ m. Figure~\ref{Fixed_CNR_RMT_NAMF_Gain} (right) compares the error (average Euclidean distance) in estimating the azimuth angle of the target with the NAMF peak cell midpoint approach (blue), a local search approach around the NAMF peak cell midpoint estimate (green), and the regression CNN approach. We observe that, in both figures, the regression CNN provides a considerable improvement in target localization accuracy, especially within the low-to-mid mean output SCNR regime, where the NAMF test statistic is close to its breakdown threshold. 

\subsection{\textbf{Evaluating Regression CNN for Variable Dataset Size} (\texorpdfstring{$\boldsymbol{N}$}{\textbf{N}})} \label{Sec7.2}

We now investigate the target localization and azimuth estimation performance of our regression CNN as we change the size, $N_{\text{train}}$, of the training dataset.  Let $\Lambda = 1, L = 16, K = 100$. We consider a mean output SCNR of $20$ dB by fixing the mean target RCS to be $\mu = M$ dBsm (with $l = 10$ dBsm) and the CNR to be $20$ dB. We vary the number of samples in the dataset, with $N$ ranging from $1 \times 10^4$ to $9 \times 10^4$, in increments of $\Delta N = 1 \times 10^4$. Again, $90\%$ of each dataset is used for training and the remaining $10\%$ is used for validation. 

\begin{figure}[h!]
    \centering
    \captionsetup{justification=centering}
    \includegraphics[width=0.95\linewidth]{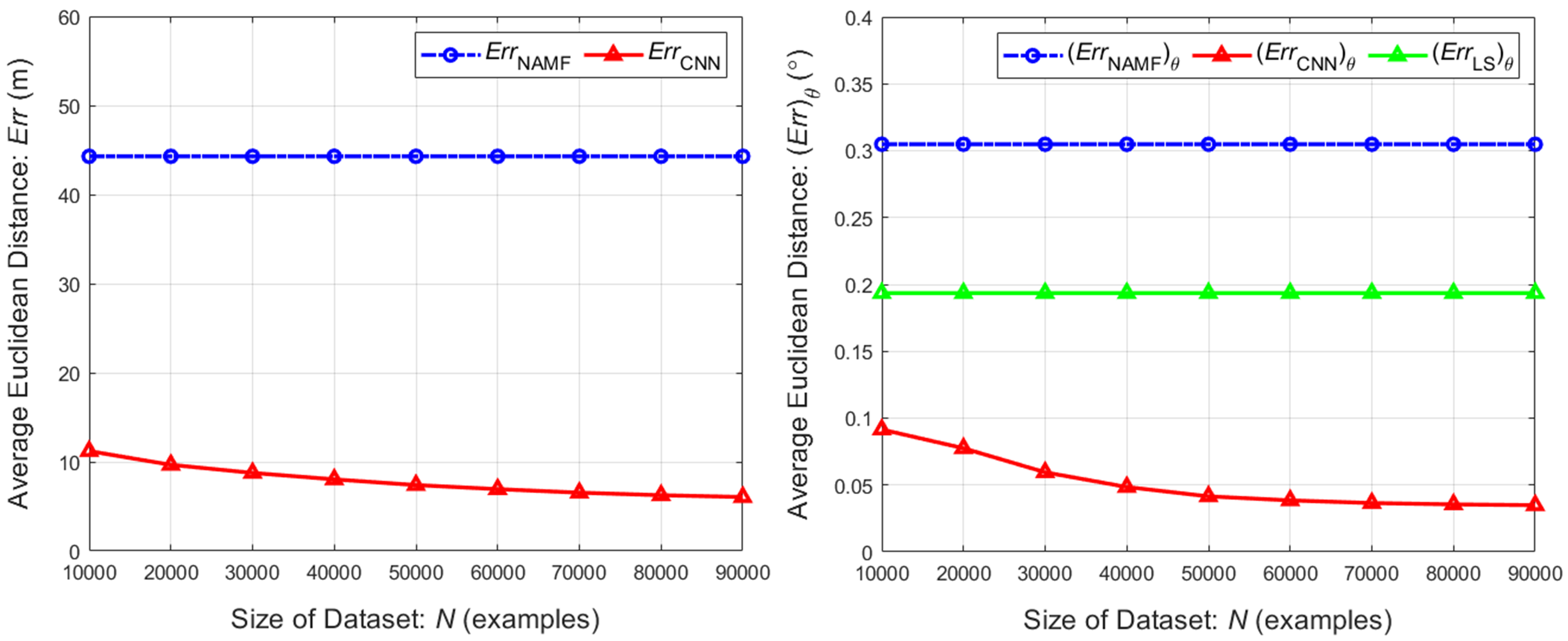}
    \caption{(Left) Target localization performance of the regression CNN (red) vs dataset size and comparison with NAMF peak cell midpoint method (blue). (Right) Azimuth estimation accuracy of the regression CNN (red) vs dataset size and comparison with azimuth estimation using NAMF peak cell midpoint (blue) and local search around the NAMF peak cell midpoint (green).}
    \label{Gain_vs_samples_NAMF}
\end{figure}

Figure~\ref{Gain_vs_samples_NAMF} (left) depicts the target localization accuracy of our regression CNN (red curve) as the size, $N$, of the dataset (and therefore, the size, $N_{\text{train}}=0.9N$, of the training dataset) increases. The blue curve depicts the target localization accuracy of the NAMF peak cell midpoint approach, which is independent of the size of the training dataset, and is provided for ease of comparison. Figure \ref{Gain_vs_samples_NAMF} (right) depicts the same comparison, specifically for estimating the azimuth angle of the target. In this plot, we have also provided the azimuth estimation accuracy obtained from a local search approach around the NAMF peak cell midpoint estimate (green curve). While increasing the dataset size improves the localization performance of the regression CNN, we observe that the gain starts to plateau around $N = 60000$ samples.

\section{\textbf{EMPIRICAL RESULTS FOR MISMATCHED CASE}} \label{Sec8}

We now investigate how the performance of our regression CNN is affected when there exists a mismatch between the radar scenario used to generate the training dataset and the radar scenario on which the network is tested. Our goal is to investigate to what extent the performance gains afforded by our CNN persist in the presence of mismatches, if at all. The mismatched scenarios are created by displacing the radar platform, such that the radar looks at different range-azimuth areas in comparison to the range-azimuth area used to generate the training dataset. This is done as described in Section~\ref{Sec3.3}. We then propose a few-shot learning approach to reduce the CNN's sensitivity to mismatches using a small number of samples from each displaced radar scenario.

\subsection{\textbf{Evaluating Regression CNN for Displaced Radar Scenarios}} \label{Sec8.2}

For our analysis, we consider the radar scenarios outlined in Section \ref{Sec3.3}. Let $\Lambda = 1, L = 16, K = 100$. We consider a mean output SCNR of $20$ dB by fixing the mean target RCS to be $\mu = M$ dBsm ($l = 10$ dBsm) and the CNR to be $20$ dB. For the original radar scenario, \textbf{(O)}, we obtain a dataset consisting of $N = 9 \times 10^4$ heatmap matrix samples. We train the regression CNN on $90\%$ of this dataset ($N_{train} = 8.1 \times 10^4$), and use the remaining $10\%$ of the dataset for validation ($N_{val} = 9 \times 10^3$). This serves as the performance benchmark for our sensitivity analysis. Next, for each of our displaced radar scenarios, \textbf{(N)}, \textbf{(W)}, \textbf{(S)}, \textbf{(E)}, we produce a dataset consisting of $0.1N$ heatmap matrix samples. These datasets are solely used to test the CNN that has been trained on the original radar scenario. Before feeding each test heatmap matrix sample (from the displaced radar scenarios), to our regression CNN trained on the original radar scenario, we perform a coordinate transformation of the heatmap matrix that translates the coordinates of the test dataset samples to those of the training dataset samples.
\begin{figure*}[h!]
    \centering
    \captionsetup{justification=centering}
    \includegraphics[width=0.95\linewidth]{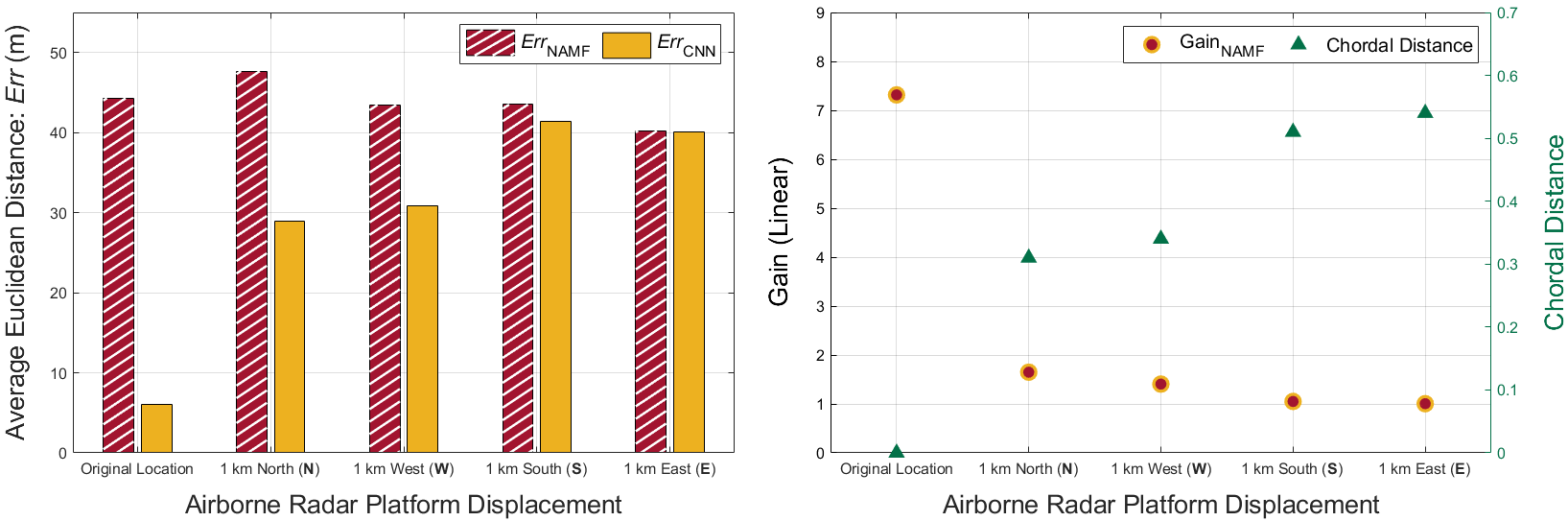}
    \caption{(Left) Target localization performance of the regression CNN (orange) when trained on the original radar scenario and tested on each displaced radar scenario, and comparison with NAMF peak cell midpoint method (red). (Right) Comparing CNN gain factor with the chordal distance between the clutter subspaces of the original and displaced radar scenarios.}
    \label{Displaced_Scenario_NAMF}
\end{figure*}
\begin{figure*}[h!]
    \centering
    \captionsetup{justification=centering}
    \includegraphics[width=0.95\linewidth]{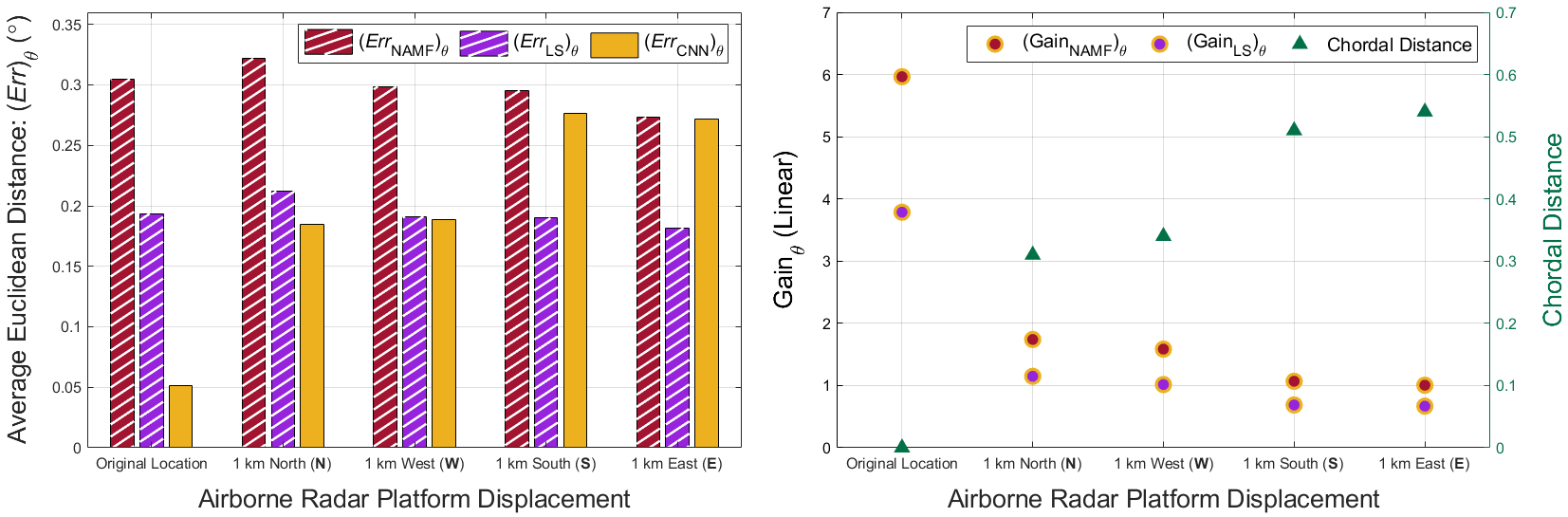}
    \caption{(Left) Azimuth estimation accuracy of the regression CNN (orange) when trained on the original radar scenario and tested on each displaced radar scenario, and comparison with NAMF peak cell midpoint method (red), and local search around the NAMF peak cell midpoint estimate (purple). (Right) Comparing CNN gain factor with the chordal distance between the clutter subspaces of the original and displaced radar scenarios.}
    \label{Displaced_Scenario_NAMF_Azimuth}
\end{figure*}

Figure~\ref{Displaced_Scenario_NAMF} (left) shows the target localization performance of our regression CNN (orange bars) for the original radar scenario, \textbf{(O)} and each of the displaced radar scenarios, \textbf{(N)}, \textbf{(W)}, \textbf{(S)}, and \textbf{(E)}. The red bars depict the localization performance of the NAMF peak cell midpoint approach, which is invariant to displacing the radar scenario. This is because we construct the NAMF test statistic anew using samples from each corresponding scenario. 

While the performance gain afforded by our regression CNN is diminished in the presence of scenario mismatches, we observe that the CNN still offers some improvement over the NAMF peak cell midpoint approach. A question that arises is whether this decline in the performance of the regression CNN is consistent with the extent of the mismatch between the radar scenarios. We investigate this by calculating the chordal distance between the clutter subspace of the original radar scenario and the clutter subspace of each displaced radar scenario --- to note, the chordal distance between two subspaces is defined as the sum of the sine squared of the principal angles between the subspaces. The blue dots in Figure~\ref{Displaced_Scenario_NAMF} (right) depict the ratio of the localization error from the NAMF peak cell midpoint approach to that of the regression CNN for each displaced scenario; that is, the gain factor afforded by the CNN. The red dots (which have their own vertical axis on the right side) depict the chordal distances between the clutter subspaces. We observe that larger performance degradations correspond to greater chordal distances. Similar trends can be observed for the azimuth estimation results provided in Figure~\ref{Displaced_Scenario_NAMF_Azimuth}. However, the CNN is no longer able to outperform a local search around the NAMF peak cell midpoint estimate. This is because the NAMF has been adapted to each displaced scenario and has no mismatch.

\subsection{\textbf{Few-Shot Learning for Displaced Radar Scenarios}} \label{Sec8.3}

We now propose an approach to make our regression CNN robust to mismatches between the radar scenario used to generate the training dataset and the radar scenario on which the network is tested. This approach is based on few-shot learning (FSL) \cite{fei_fei_one_shot, vinyals2016matching, ravi2017optimization, sung2018learning} --- a method of training machine learning algorithms using minimal data.

FSL is a machine learning method that builds on the notion of one-shot learning: a challenging task where a single example is used to train a learning model to classify objects (see \cite{fei_fei_one_shot,fink_single_example}). Over the past decade, one-shot learning has become foundational to the problem of visual object detection, which is a computer vision paradigm pertaining to the identification and localization of objects in images \& videos. In particular, single-shot detectors such as YOLO \cite{redmon2016you} and YOLOv3 \cite{redmon_yolov3} achieve near real-time visual object detection, depicting the feasibility of one-shot learning in computer vision applications. As an extension of these methods, FSL has more recently been used in conjunction with transfer learning and fine tuning via a data-level approach \cite{shen_partial}. By training the provided learning model on a large base dataset for a specified task, we can fine-tune the learning model to perform a similar task using minimal new examples via FSL.

In our few-shot learning approach, we first train the regression CNN from Figure~\ref{CNN_default_parameters} on the original radar scenario, \textbf{(O)}, using $N_{train} = 8.1 \times 10^4$ heatmap matrix samples (see Section \ref{Sec8.2} for all relevant parameter values), and then fine-tune the CNN with a small number of samples from the new (displaced) radar scenario --- we freeze the convolutional and batch normalization layers of our trained regression CNN (the weights and biases will remain constant), which halves the number of trainable parameters to $6{\small,}462$. Subsequently, we generate a dataset consisting of $64$ new heatmap matrices for each displaced radar scenario. Using these $64$ new heatmap matrices, the weights and biases of the remaining layers will be updated via fine-tuning with FSL. 

\begin{figure*}[h!]
    \centering
    \captionsetup{justification=centering}
    \includegraphics[width=0.95\linewidth]{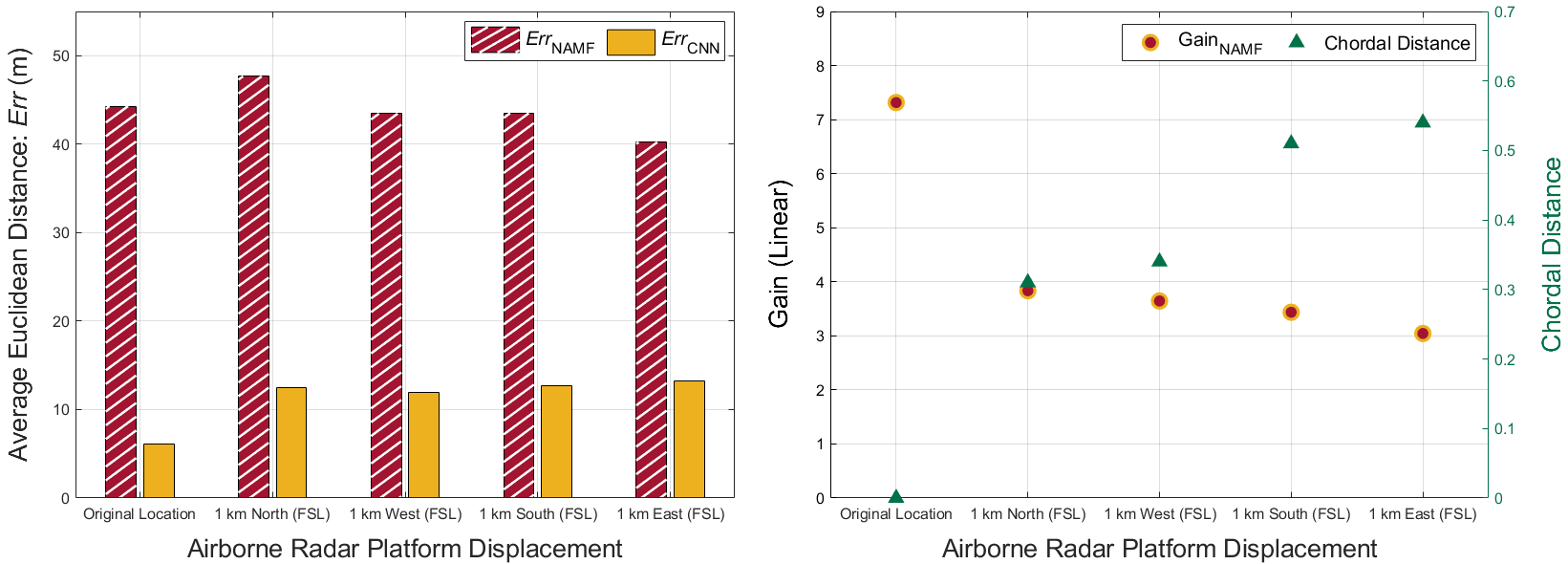}
    \caption{(Left) Target localization performance of the regression CNN (orange) when trained on the original radar scenario and tested on each displaced radar scenario after fine tuning using FSL, and comparison with NAMF peak cell midpoint method (red). (Right) Comparing CNN gain factor after fine tuning using FSL with the chordal distance between the clutter subspaces of the original and displaced radar scenarios.}
    \label{Displaced_Scenario_NAMF_FSL}
\end{figure*}

\begin{figure*}[h!]
    \centering
    \captionsetup{justification=centering}
    \includegraphics[width=0.95\linewidth]{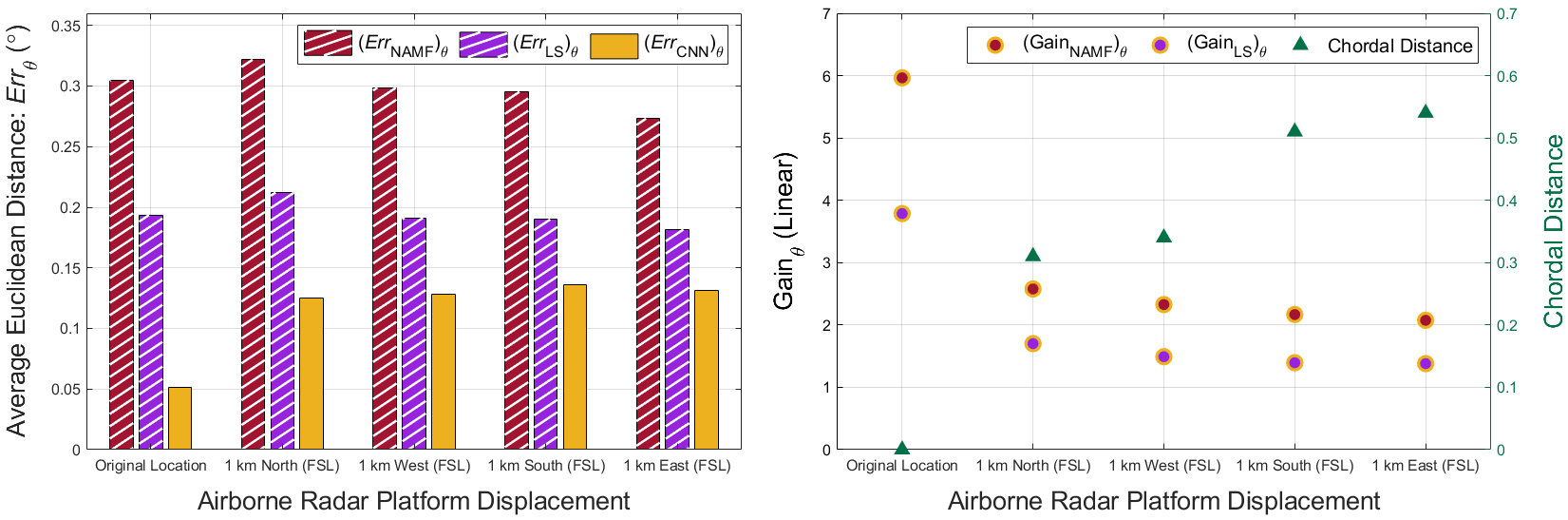}
    \caption{(Left) Azimuth estimation accuracy of the regression CNN (orange) when trained on the original radar scenario and tested on each displaced radar scenario after fine tuning using FSL, and comparison with NAMF peak cell midpoint method (red), and local search around the NAMF peak cell midpoint estimate (purple). (Right) Comparing CNN gain factor after fine tuning using FSL with the chordal distance between the clutter subspaces of the original and displaced radar scenarios.}
    \label{Displaced_Scenario_NAMF_FSL_Azimuth}
\end{figure*}

Figure~\ref{Displaced_Scenario_NAMF_FSL} (left) shows the target localization performance of our regression CNN (orange bars) for the original radar scenario, \textbf{(O)}, and for each of the displaced radar scenarios, \textbf{(N)}, \textbf{(W)}, \textbf{(S)}, \textbf{(E)}, after fine-tuning with FSL. The red bars depict the localization performance of the NAMF peak cell midpoint approach. Similarly, Figure~\ref{Displaced_Scenario_NAMF_FSL_Azimuth} (left) depicts the azimuth estimation accuracy of our regression CNN (orange bars) for the original radar scenario, and for each of the displaced radar scenarios after fine-tuning with FSL. The red and purple bars depict the azimuth estimation accuracy of the NAMF peak cell midpoint approach, and the local search approach around the NAMF peak cell midpoint estimate, respectively. Across both figures, we observe that the gains afforded by the CNN over the NAMF peak cell midpoint approach, and the local search around the NAMF peak cell midpoint estimate have been largely restored after fine tuning with FSL.

% We now conclude our discussion by making the following distinction. While the performance of our regression CNN framework is a monotonic function of $(\overline{\text{SCNR}}_{\text{Normalized Output}})_{\hat{\rho}}$, deriving the exact relationship quickly becomes mathematically intractable due to the black-box nature of the CNN. As this translates to our few-shot learning evaluation, determining the exact improvement in $(\overline{\text{SCNR}}_{\text{Normalized Output}})_{\hat{\rho}}$ for a given displaced platform location instance --- after FSL has been applied to it --- is nontrivial. Such an investigation exists outside the scope of our analysis.

\section{\textbf{EMPIRICAL RESULTS FOR DOPPLER CASE}} \label{Sec9}
Finally, in order to assess the capability of our Doppler CNN from Figure~\ref{CNN_Doppler} in estimating both target location and velocity, we examine an example scenario involving Doppler processing, as discussed in Section \ref{Sec3.4}. There is no radar scenario mismatch in this analysis, and we evaluate the performance of the Doppler CNN on the same radar scenario on which it is trained, although using a separate validation dataset. Let $\Lambda = 4, L = 16, K = 400$. We vary the mean output SCNR from $-20$ dB to $20$ dB by changing the mean target RCS, $\mu$ (with $l = 10$ dBsm), for a fixed CNR of $20$ dB. For each mean output SCNR value, we obtain a dataset consisting of $N = 9 \times 10^4$ heatmap matrix samples. We utilize $90\%$ of the dataset for training ($N_{train} = 8.1 \times 10^4$) and the remaining $10\%$ for validation ($N_{val} = 9 \times 10^3$). We train the Doppler CNN on the training dataset using the Adam optimizer~\cite{kingma_14}, with learning rate hyperparameter $\alpha = 1 \times 10^{-3}$, and weight decay hyperparameter $\lambda = 1 \times 10^{-3}$.

\begin{figure*}[h!]
    \centering
    \captionsetup{justification=centering}
    \includegraphics[width=0.95\linewidth]{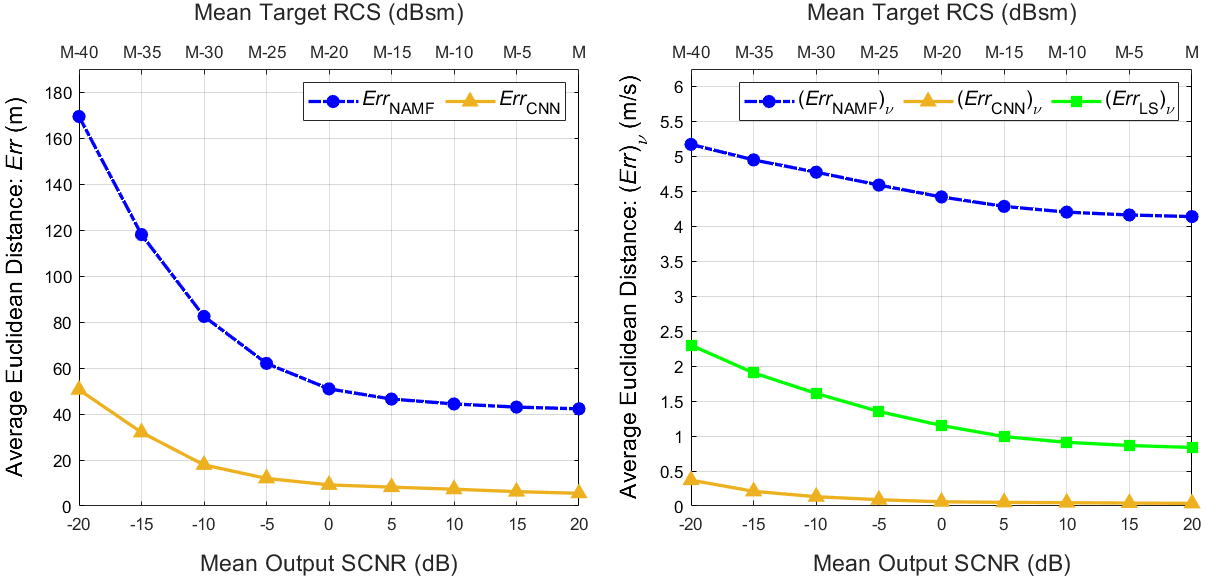}
    \caption{(Left) Target localization performance of the regression CNN (orange) vs SCNR and comparison with NAMF peak cell midpoint method (blue). (Right) Velocity estimation accuracy of the regression CNN (orange) vs SCNR and comparison with velocity estimation using NAMF peak cell midpoint (blue) and local search around the NAMF peak cell midpoint (green).}
    \label{Fixed_CNR_Doppler_Gain}
\end{figure*}

Figure~\ref{Fixed_CNR_Doppler_Gain} (left) compares the target localization accuracy of the Doppler CNN with that of the NAMF peak cell midpoint method. Figure~\ref{Fixed_CNR_Doppler_Gain} (right) compares the velocity estimation accuracy of the Doppler CNN with that of standard Doppler processing using the NAMF peak cell midpoint approach and the local search approach around the NAMF peak cell midpoint estimate. In both cases, the regression CNN yields considerable performance gains.

\section{\textbf{CONCLUDING REMARKS}} \label{Sec10}
The performance of covariance-based and subspace-based radar estimation approaches suffers considerable degradation as the SCNR approaches the breakdown threshold predicted by random matrix theory and subspace swap analysis methods. A natural question to ask is whether post-processing the test statistics obtained from such methods with deep neural networks can restore, or even improve estimation performance. Another question is where we can obtain sufficient data for training such neural networks for post-processing. Furthermore, are such networks robust to mismatches between the training dataset and the test dataset?

In this paper, we empirically studied the above questions. The emergence of RF simulators that leverage topographic and material composition datasets for high-fidelity radar simulations solves the data paucity problem. We leveraged the RFView\textsuperscript{\tiny\textregistered} simulator, which has been cross-validated against real radar data, for this purpose. We demonstrated that we can train regression convolutional neural networks for target localization using heatmap tensors of NAMF test statistics. Our empirical results showed that these trained CNNs offer considerable performance improvements relative to peak-finding and local search (around the peak) approaches with NAMF processing alone. We also investigated the sensitivity of our CNNs to mismatches caused by changing the radar processing region, and proposed a few-shot learning method to enhance robustness.

% \ap{The performance of covariance and subspace-based radar estimation methods is likely to suffer considerable degradation as SCNR nears breakdown threshold predicted by random matrix theory and subspace swap analysis methods. A natural question to ask is whether post processing the test statistics obtained form such methods with deep neural networks can restore or even improve estimation performance? Another question is where can be obtain sufficient data for training such neural networks for post processing? And finally, are such networks robust to mismatch between the training data set and the testing data set?

% In this paper, we empirically studied the above questions. The emergence of RF simulators that leverage topographic and material compositions data sets for high fidelity radar simulations solves the data paucity problem. We leveraged the RFView simulator, which has been cross-validated against real radar data, for this purpose. We showed we can train regression CNN networks for target localization from heatmap tensors of NAMF test statistics. 
% Our empirical results showed that the trained network offer considerable performance improvement relative to peak finding and local search (around the peak) approaches with NAMF processing only. We also investigated the sensitivity of our regression CNNs to mismatch caused by changing the platform location and proposed a few-shot learning method for robustyfying it.}

\section*{\textbf{ACKNOWLEDGMENT}}
This work is supported in part by the Air Force Office of Scientific Research (AFOSR) under award FA9550-21-1-0235. Dr. Muralidhar Rangaswamy and Dr. Bosung Kang are supported by the AFOSR under project 20RYCORO51. Dr. Sandeep Gogineni is supported by the AFOSR under project 20RYCOR052. The opinions and statements within this paper are the authors’ own and do not constitute any explicit or implicit endorsement by the U.S. Department of Defense.

\end{spacing}

\bibliographystyle{IEEEtran}
\bibliography{references}

\begin{thebibliography}{10}
\providecommand{\url}[1]{#1}
\csname url@samestyle\endcsname
\providecommand{\newblock}{\relax}
\providecommand{\bibinfo}[2]{#2}
\providecommand{\BIBentrySTDinterwordspacing}{\spaceskip=0pt\relax}
\providecommand{\BIBentryALTinterwordstretchfactor}{4}
\providecommand{\BIBentryALTinterwordspacing}{\spaceskip=\fontdimen2\font plus
\BIBentryALTinterwordstretchfactor\fontdimen3\font minus \fontdimen4\font\relax}
\providecommand{\BIBforeignlanguage}[2]{{%
\expandafter\ifx\csname l@#1\endcsname\relax
\typeout{** WARNING: IEEEtran.bst: No hyphenation pattern has been}%
\typeout{** loaded for the language `#1'. Using the pattern for}%
\typeout{** the default language instead.}%
\else
\language=\csname l@#1\endcsname
\fi
#2}}
\providecommand{\BIBdecl}{\relax}
\BIBdecl

\bibitem{kraut_adaptive}
S.~Kraut, L.~Scharf, and L.~McWhorter, ``Adaptive subspace detectors,'' \emph{IEEE Transactions on Signal Processing}, vol.~49, no.~1, pp. 1--16, 2001.

\bibitem{melvin_stap96}
W.~Melvin, M.~Wicks, and R.~Brown, ``Assessment of multichannel airborne radar measurements for analysis and design of space-time processing architectures and algorithms,'' in \emph{Proceedings of the 1996 IEEE National Radar Conference}, 1996, pp. 130--135.

\bibitem{guerci2003space}
J.~Guerci, \emph{Space-time Adaptive Processing for Radar}, ser. Artech House radar library.\hskip 1em plus 0.5em minus 0.4em\relax Artech House, 2003.

\bibitem{guerci_stap00}
J.~Guerci, J.~Goldstein, and I.~Reed, ``Optimal and adaptive reduced-rank stap,'' \emph{IEEE Transactions on Aerospace and Electronic Systems}, vol.~36, no.~2, pp. 647--663, 2000.

\bibitem{raghavan_CFAR}
R.~Raghavan, H.~Qiu, and D.~McLaughlin, ``Cfar detection in clutter with unknown correlation properties,'' \emph{IEEE Transactions on Aerospace and Electronic Systems}, vol.~31, no.~2, pp. 647--657, 1995.

\bibitem{gogineni_RFView}
S.~Gogineni, J.~R. Guerci, H.~K. Nguyen, J.~S. Bergin, D.~R. Kirk, B.~C. Watson, and M.~Rangaswamy, ``High fidelity rf clutter modeling and simulation,'' \emph{IEEE Aerospace and Electronic Systems Magazine}, vol.~37, no.~11, pp. 24--43, 2022.

\bibitem{krizhevsky2009learning}
A.~Krizhevsky, G.~Hinton \emph{et~al.}, ``Learning multiple layers of features from tiny images,'' \emph{Citeseer}, 2009.

\bibitem{ILSVRC15}
O.~Russakovsky, J.~Deng, H.~Su, J.~Krause, S.~Satheesh, S.~Ma, Z.~Huang, A.~Karpathy, A.~Khosla, M.~Bernstein, A.~C. Berg, and L.~Fei-Fei, ``{ImageNet Large Scale Visual Recognition Challenge},'' \emph{International Journal of Computer Vision (IJCV)}, vol. 115, no.~3, pp. 211--252, 2015.

\bibitem{Benaych_eigenvalues}
F.~Benaych-Georges and R.~R. Nadakuditi, ``The eigenvalues and eigenvectors of finite, low rank perturbations of large random matrices,'' \emph{Advances in Mathematics}, vol. 227, no.~1, pp. 494--521, 2011.

\bibitem{Gogineni_Passive}
S.~Gogineni, P.~Setlur, M.~Rangaswamy, and R.~R. Nadakuditi, ``Passive radar detection with noisy reference channel using principal subspace similarity,'' \emph{IEEE Transactions on Aerospace and Electronic Systems}, vol.~54, no.~1, pp. 18--36, 2018.

\bibitem{Nadakuditi_correlation}
R.~R. Nadakuditi, ``Fundamental finite-sample limit of canonical correlation analysis based detection of correlated high-dimensional signals in white noise,'' in \emph{2011 IEEE Statistical Signal Processing Workshop (SSP)}, 2011, pp. 397--400.

\bibitem{rangaswamy_NAMF}
M.~Rangaswamy and F.~Lin, ``Normalized adaptive matched filter - a low rank approach,'' in \emph{Proceedings of the 3rd IEEE International Symposium on Signal Processing and Information Technology (IEEE Cat. No.03EX795)}, 2003, pp. 182--185.

\bibitem{conteNAMF}
E.~Conte, ``\BIBforeignlanguage{English}{Multistatic radar detection: synthesis and comparison of optimum and suboptimum receivers},'' \emph{\BIBforeignlanguage{English}{IEE Proceedings F (Communications, Radar and Signal Processing)}}, vol. 130, pp. 484--494(10), October 1983.

\bibitem{scharf_adaptive}
L.~Scharf and L.~McWhorter, ``Adaptive matched subspace detectors and adaptive coherence estimators,'' in \emph{Conference Record of The Thirtieth Asilomar Conference on Signals, Systems and Computers}, 1996, pp. 1114--1117 vol.2.

\bibitem{scharf1996adaptive}
L.~L. Scharf, T.~L. McWhorter, and L.~J. Griffiths, ``Adaptive coherence estimation for radar signal processing,'' in \emph{Proceedings of the 30th Asilomar Conference on Signals, Systems, and Computers}, Pacific Grove, CA, 1996.

\bibitem{kraut1997canonical}
S.~Kraut, T.~L. McWhorter, and L.~L. Scharf, ``A canonical representation for the distributions of adaptive matched subspace detectors,'' in \emph{Proceedings of the 31st Asilomar Conference on Signals, Systems, and Computers}, Pacific Grove, CA, 1997.

\bibitem{Shyam_STAP}
S.~Venkatasubramanian, C.~Wongkamthong, M.~Soltani, B.~Kang, S.~Gogineni, A.~Pezeshki, M.~Rangaswamy, and V.~Tarokh, ``Toward data-driven stap radar,'' in \emph{2022 IEEE Radar Conference (RadarConf22)}, 2022, pp. 1--5.

\bibitem{capon_mvdr}
J.~Capon, ``High-resolution frequency-wavenumber spectrum analysis,'' \emph{Proceedings of the IEEE}, vol.~57, no.~8, pp. 1408--1418, 1969.

\bibitem{lecun1998gradient}
Y.~Lecun, L.~Bottou, Y.~Bengio, and P.~Haffner, ``Gradient-based learning applied to document recognition,'' \emph{Proceedings of the IEEE}, vol.~86, no.~11, pp. 2278--2324, 1998.

\bibitem{tufts1991threshold}
D.~Tufts, A.~Kot, and R.~Vaccaro, ``The threshold effect in signal processing algorithms which use an estimated subspace,'' in \emph{SVD and Signal Processing II: Algorithms, Analysis and Applications}.\hskip 1em plus 0.5em minus 0.4em\relax New York, NY, USA: Elsevier, 1991, pp. 301--320.

\bibitem{thomas1995probability}
J.~K. Thomas, L.~L. Scharf, and D.~W. Tufts, ``The probability of a subspace swap in the svd,'' \emph{IEEE Trans. Signal Process.}, vol.~43, no.~3, pp. 730--736, 1995.

\bibitem{pakrooh2016threshold}
P.~Pakrooh, L.~L. Scharf, and A.~Pezeshki, ``Threshold effects in parameter estimation from compressed data,'' \emph{IEEE Trans. Signal Processing}, vol.~64, no.~9, pp. 2345--2354, May 2016.

\bibitem{kingma_14}
\BIBentryALTinterwordspacing
D.~P. Kingma and J.~Ba, ``Adam: A method for stochastic optimization,'' in \emph{International Conference for Learning Representations}, 2015. [Online]. Available: \url{http://arxiv.org/abs/1412.6980}
\BIBentrySTDinterwordspacing

\bibitem{fei_fei_one_shot}
L.~Fei-Fei, R.~Fergus, and P.~Perona, ``One-shot learning of object categories,'' \emph{IEEE Transactions on Pattern Analysis and Machine Intelligence}, vol.~28, no.~4, pp. 594--611, 2006.

\bibitem{vinyals2016matching}
O.~Vinyals, C.~Blundell, T.~Lillicrap, K.~Kavukcuoglu, and D.~Wierstra, ``Matching networks for one shot learning,'' in \emph{Advances in Neural Information Processing Systems (NeurIPS)}, 2016, pp. 3630--3638.

\bibitem{ravi2017optimization}
S.~Ravi and H.~Larochelle, ``Optimization as a model for few-shot learning,'' in \emph{Proceedings of the International Conference on Learning Representations (ICLR)}, 2017.

\bibitem{sung2018learning}
F.~Sung, Y.~Yang, L.~Zhang, T.~Xiang, P.~H. Torr, and T.~M. Hospedales, ``Learning to compare: Relation network for few-shot learning,'' in \emph{Proceedings of the IEEE Conference on Computer Vision and Pattern Recognition (CVPR)}, 2018, pp. 1199--1208.

\bibitem{fink_single_example}
M.~Fink, ``Object classification from a single example utilizing class relevance metrics,'' in \emph{Advances in Neural Information Processing Systems}, L.~Saul, Y.~Weiss, and L.~Bottou, Eds., vol.~17.\hskip 1em plus 0.5em minus 0.4em\relax MIT Press, 2004.

\bibitem{redmon2016you}
J.~Redmon, S.~Divvala, R.~Girshick, and A.~Farhadi, ``You only look once: Unified, real-time object detection,'' in \emph{Proceedings of the IEEE conference on computer vision and pattern recognition}, 2016, pp. 779--788.

\bibitem{redmon_yolov3}
J.~Redmon and A.~Farhadi, ``Yolov3: An incremental improvement,'' 2018.

\bibitem{shen_partial}
Z.~Shen, Z.~Liu, J.~Qin, M.~Savvides, and K.-T. Cheng, ``Partial is better than all: Revisiting fine-tuning strategy for few-shot learning,'' 2021.

\end{thebibliography}

\clearpage
\end{document}